%% file: acl_latex.tex
\documentclass[11pt]{article}

\usepackage[final]{acl}

\usepackage{times}
\usepackage{latexsym}

\usepackage[T1]{fontenc}

\usepackage[utf8]{inputenc}

\usepackage{microtype}

\usepackage{inconsolata}

\usepackage{booktabs}
\usepackage{graphicx}
\usepackage{amssymb}
\usepackage{amsmath} 
\usepackage{arydshln} 
\usepackage{multirow}
\usepackage{bm} 

\newcommand{\accstd}[2]{\ensuremath{#1_{\scriptscriptstyle #2}}}
\newcommand\blfootnote[1]{%
  \begingroup
  \renewcommand\thefootnote{}\footnote{#1}%
  \addtocounter{footnote}{-1}%
  \endgroup
}

\title{When In-Distribution Gains Fail: Evaluating Weak-to-Strong Reward Models under Preference Shift}

\author{Khoi Le$^{\dagger *\,1}$ \quad Tri Cao$^{*\,1}$ \quad Phong Nguyen $^2$ \quad Cong-Duy Nguyen $^2$, \\ \textbf{Anh-Tuan Luu $^{3,2}$ \quad Miao Chunyan $^3$ \quad See-Kiong Ng $^1$ \quad Thong Nguyen $^{1}$}
\\\\
$^{1}$ National University of Singapore, $^{2}$ VinUniversity, $^{3}$ Nanyang Technological University
}

\begin{document}
\maketitle
\begin{abstract}
Weak-to-strong (W2S) generalization is a promising framework for scalable oversight, yet existing evaluations often test students under matched train--test distributions. Therefore, we study W2S preference learning under zero-shot distribution shift and find that strong students trained on weak preference labels can appear successful in-distribution while failing to transfer across preference datasets. We provide evidence consistent with a representational failure mode in which weak-supervised fine-tuning may move the strong model toward source-domain features instead of maintaining broadly transferable preference representations. To mitigate this, we propose \emph{Representation Anchoring} (\textsc{Anchor}), a simple yet effective regularizer that constrains excessive drift from the pretrained strong model's representation space during fine-tuning, while still allowing task-relevant adaptation. Across preference domains, datasets, and model families, \textsc{Anchor} improves out-of-distribution transfer while maintaining competitive in-distribution performance. Together, our evaluation protocol, transfer-aware metrics, and method expose hidden brittleness in current W2S reward modeling and provide a practical path toward more robust preference transfer.
\footnote{Code: \url{https://github.com/TokisakiKurumi2001/w2s_ood}}
\blfootnote{$^{*}$Contributes equally, $^{\dagger}$Corresponding author.}
\blfootnote{Email: minhkhoi.le@u.nus.edu}
\end{abstract}

\input{content/01_intro}
\input{content/02_preliminary}
\input{content/03_problem_formulation}
\input{content/04_method}
\input{content/05_exp}
\input{content/06_analysis}
\input{content/07_related_work}
\input{content/08_conclusion}
\input{content/09_limitation}
\input{content/10_ack}

\bibliography{custom}

\clearpage
\appendix

\input{content/appendix}

\end{document}

%% file: content/01_intro.tex
\section{Introduction}

\begin{figure}[t]
    \centering
    \includegraphics[width=\linewidth]{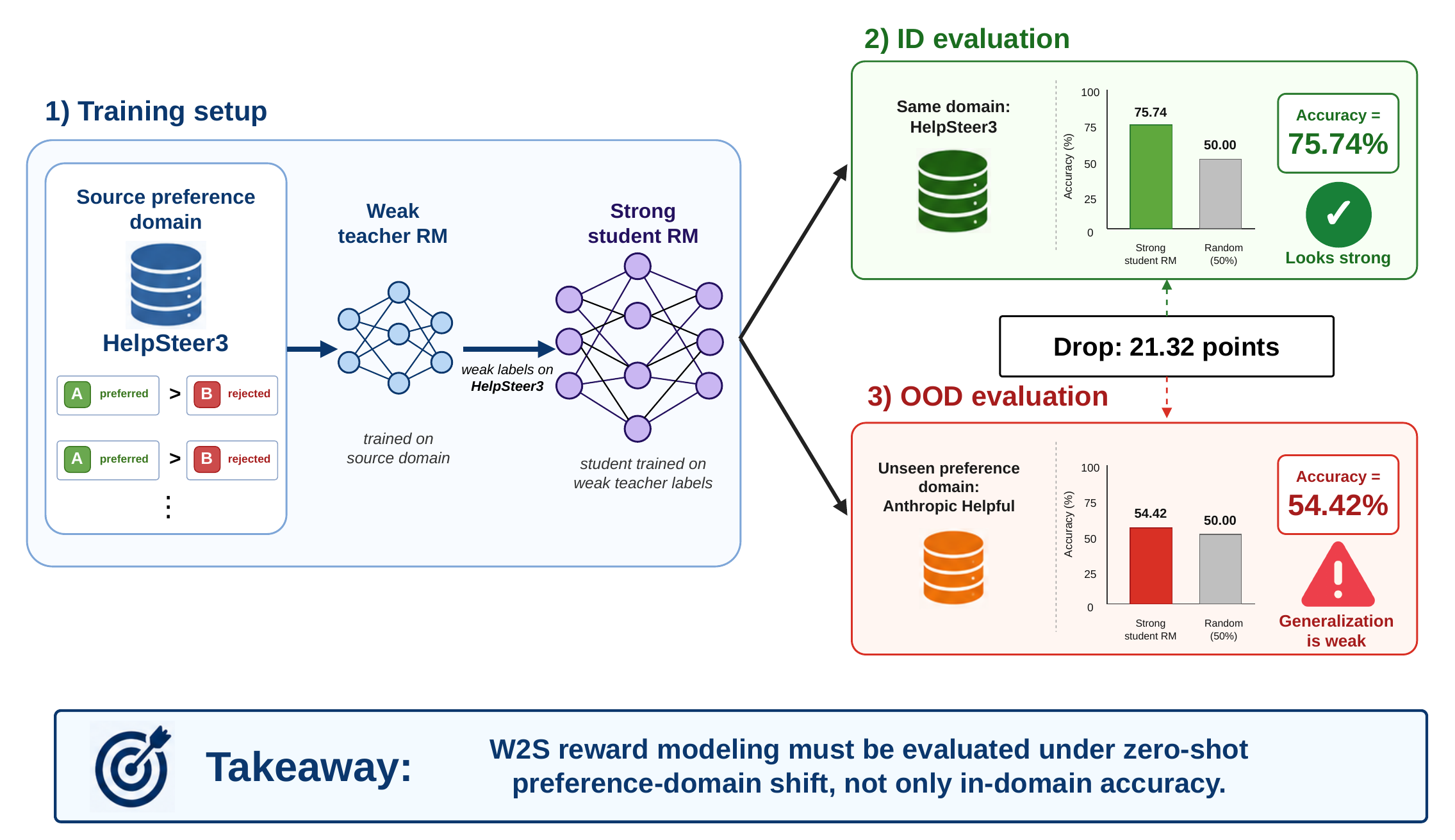}
    \caption{\textbf{ID success can hide OOD failure.} Although the resulting student can perform well on in-domain evaluation, its performance can degrade substantially when tested on unseen preference domains. This motivates evaluating W2S reward models under zero-shot preference-domain shift rather than relying only on in-domain accuracy.}
    \label{fig:intro}
\end{figure}

As AI systems become increasingly capable, a central challenge is how to provide reliable supervision when humans cannot consistently evaluate model outputs due to limited expertise, cost, or scale. This challenge has motivated work on scalable oversight, which studies how weaker supervisors can guide stronger systems when direct supervision is costly, unreliable, or incomplete \citep{amodei2016concreteproblemsaisafety,leike2018scalableagentalignmentreward,bowman2022measuringprogressscalableoversight}. Weak-to-strong generalization (W2S) provides a concrete abstraction of this problem: a stronger student is trained on labels produced by a weaker teacher, with the goal of eliciting capabilities beyond the teacher's own performance \citep{burns2023weakstrong}.

Recent W2S methods have shown encouraging results across reasoning, mathematical problem solving, instruction tuning, and preference optimization \citep{burns2023weakstrong,yang2024weak,li2024superfiltering,zhu2025weak}. These findings suggest that strong students can often extract useful task structure from imperfect weak supervision, enabling them to outperform their weak teachers rather than simply inherit the teacher's limitations.

However, most existing W2S evaluations still test students under matched or near-matched train-test distributions, which may overestimate alignment reliability in realistic deployment. We instead study preference-domain shift: the weak supervisor and student are trained on one preference dataset and evaluated zero-shot on another dataset within the same broad preference category, where prompts, response styles, annotation protocols, and preference policies may differ. Unlike an ordinary held-out split, this setting tests whether weak-supervised reward models learn transferable preference criteria rather than dataset-specific patterns. This gap motivates our central question: \emph{does weak-to-strong preference learning transfer alignment behavior beyond the distribution from which weak supervision is generated?}

To answer this question, we introduce a zero-shot distribution-shift evaluation protocol for W2S, focusing on reward modeling as a central component of preference-based alignment pipelines \citep{stiennon2020learning,ouyang2022training,bai2022training}. Within each broad preference domain, such as helpfulness or harmlessness, models are trained on one dataset and evaluated both on its held-out test split and on other datasets from the same domain but with different annotation sources and preference styles. Across domains, datasets, and model families, we find that existing W2S methods often match or approach strong ground-truth performance in-distribution, yet remain fragile under cross-distribution evaluation. As shown in Figure~\ref{fig:intro}, a reward model trained on HelpSteer3 \citep{wang2025helpsteer3preference} achieves strong in-distribution performance, but its accuracy drops substantially when evaluated on Anthropic Helpful \citep{bai2022training}.

To improve robustness under this shift, we propose \emph{Representation Anchoring} (\textsc{Anchor}), which constrains excessive drift from the pretrained strong model’s representation space during fine-tuning. \textsc{Anchor} is designed to limit excessive distortion of broadly useful pretrained features while still allowing the reward model to adapt to weak preference labels. Across preference domains and model families, \textsc{Anchor} achieves the most balanced overall performance over existing W2S methods under both in-distribution and out-of-distribution evaluation.

In summary, our contributions are threefold:
\begin{itemize}
    \vspace{-2mm}
    \item We introduce a zero-shot preference-domain shift protocol for evaluating W2S reward modeling across datasets, domains, and model families.

    \vspace{-2mm}
    \item Using this protocol, we show that existing W2S methods can appear successful in-distribution yet remain fragile under cross-domain evaluation, revealing that in-distribution performance can overestimate alignment reliability.
    
    \item We propose \emph{Representation Anchoring} (\textsc{Anchor}), a simple yet effective method that regularizes the student toward the pretrained strong model's representations, limiting excessive representation drift during weak-supervised training. Across empirical experiments, \textsc{Anchor} yields the most balanced gains under both in-distribution and out-of-distribution evaluation.
\end{itemize}

%% file: content/02_preliminary.tex
\section{Preliminaries}
\label{sec:preliminary}

\subsection{Weak-to-Strong Generalization}
\label{sec:w2s_generalization}

Weak-to-strong (W2S) generalization studies whether a stronger model can be trained from supervision produced by a weaker model \citep{burns2023weakstrong}. In each source domain \(S\), we instantiate three models: a weak supervisor \(w_S\), a strong student \(m_S\), and a strong ceiling model \(m^{\mathrm{gt}}_S\). The weak supervisor is trained on gold labels and then used to supervise the strong student. The strong ceiling model has the same base capacity as \(m_S\), but is trained directly on gold labels and is used only for evaluation.

In this work, we study W2S in the setting of preference reward modeling, where the supervised object is a scalar reward model trained from pairwise preferences.

\subsection{Preference Reward Modeling}
\label{sec:pref_rm}

A preference dataset consists of prompts paired with a preferred and rejected response,
\[
\mathcal{D}=\{(x_i,y_i^+,y_i^-)\}_{i=1}^N.
\]
A reward model \(r_\theta(x,y)\in\mathbb{R}\) assigns a scalar score to each prompt--response pair. Following the Bradley--Terry model \citep{bradley1952rank}, the probability of preferring \(y_i^+\) over \(y_i^-\) is
\[
p_\theta(y_i^+ \succ y_i^- \mid x_i)
= \sigma\!\left(r_\theta(x_i,y_i^+) - r_\theta(x_i,y_i^-)\right),
\]
and the standard reward-modeling objective is
\[
\mathcal{L}_{\mathrm{RM}}(\theta)
= -\frac{1}{N}\sum_{i=1}^{N}
\log \sigma\!\left( r_\theta(x_i,y_i^+) - r_\theta(x_i,y_i^-) \right).
\]
In supervised reward modeling, \((y_i^+,y_i^-)\) is determined by human preference labels, as in RLHF-style alignment pipelines \citep{christiano2017deep,ouyang2022training,bai2022training}. In W2S reward modeling, the strong student instead learns from preference labels induced by the weak supervisor.

\subsection{Weak-to-Strong Reward-Modeling Protocol}
\label{sec:w2s_protocol}

Given a source training set \(S_{\mathrm{train}}\), we split it into disjoint supervisor and student subsets. We train \(w_S\) on gold preference labels from the supervisor subset, and then use \(w_S\) to annotate each example in the student subset with a soft preference label
\[
q_i = p_{w_S}(y_i^+ \succ y_i^- \mid x_i).
\]
The strong student \(m_S\) is trained only on these weak labels, while the strong ceiling model \(m^{\mathrm{gt}}_S\) is trained on gold labels using the same strong model base capacity.

%% file: content/03_problem_formulation.tex
\section{Problem definition}
\label{sec:problem}

We now formalize the central question of this work: whether weak-to-strong reward modeling learns domain-general preference representations or merely fits source-domain preference patterns.

\subsection{Zero-Shot Preference-Domain Shift}
\label{sec:zero_shot_shift}

Let \(\mathcal{C}\) denote a broad preference category, such as helpfulness or harmlessness. Each category contains a collection of preference domains
\[
\mathcal{K}_{\mathcal{C}} = \{k_1, k_2, \ldots, k_M\},
\]
where each \(k_i\) corresponds to a distinct data distribution with its own prompts,
response styles, annotation protocol, and dataset artifacts. For each source domain
\(k_i \in \mathcal{K}_{\mathcal{C}}\), we train a weak-to-strong reward model
\(r_{\theta}\) using only \(k_i^{\mathrm{train}}\). We evaluate \(r_{\theta}\) in-distribution
on the held-out test set from the same domain,
\[
\mathcal{E}_{\mathrm{ID}}(i) = \{k_i^{\mathrm{test}}\},
\]
and out-of-distribution on the test sets from all remaining domains,
\[
\mathcal{E}_{\mathrm{OOD}}(i)
= \{k_j^{\mathrm{test}} : k_j \in \mathcal{K}_{\mathcal{C}},\ j \neq i\}.
\]

We focus on \emph{zero-shot preference-domain transfer}: no target-domain examples are used to train the weak supervisor, generate weak labels, or tune the reward head. Hence, target-domain performance reflects transfer from source-domain weak supervision rather than target-domain adaptation.

This formulation differs from ordinary held-out evaluation. A held-out source test set asks whether the student learned the source preference distribution, while a target-domain test set asks whether the learned preference signal remains useful when the prompt distribution, response style, or dataset artifacts change. This distinction is central to scalable oversight: weak supervision may be available only for limited behavioral slices, whereas deployed reward models must generalize across many shifted preference domains.

\subsection{Transfer-Aware Metrics}
\label{sec:metrics}

Raw preference accuracy is necessary but insufficient for evaluating weak-to-strong reward modeling under preference-domain shift. A method may improve on the source domain while failing to transfer, or improve on target domains only by degrading source-domain behavior. We therefore report three transfer-aware metrics that separately capture source-domain improvement, target-domain transfer, and transfer after accounting for source-domain regression.

Let \(S \in \mathcal{K}_{\mathcal{C}}\) be the source preference domain used to train the weak supervisor \(w_S\) and the strong student \(m_S\). For any evaluation domain \(E \in \mathcal{K}_{\mathcal{C}}\), define
\[
    A_w(S,E) = \mathrm{Acc}(w_S, E_{\mathrm{test}}),
\]
\[
    A_m(S,E) = \mathrm{Acc}(m_S, E_{\mathrm{test}}),
\]
where \(\mathrm{Acc}(\cdot, E_{\mathrm{test}})\) is pairwise preference accuracy on the test split of domain \(E\). The case \(E=S\) corresponds to in-distribution evaluation, while \(E=T\neq S\) corresponds to zero-shot out-of-distribution evaluation on a target domain \(T\).

\paragraph{Weak-to-Strong Raw Gain.}
Weak-to-Strong Raw Gain (WRG) \citep{charikar2024quantifying} measures whether the strong student improves over the weak supervisor on the source domain:
\[
    \mathrm{WRG}(S;m) = A_m(S,S) - A_w(S,S).
\]
Positive WRG indicates successful in-distribution weak-to-strong improvement.

\paragraph{Absolute OOD Gain.}
For an unseen target domain \(T \neq S\), Absolute OOD Gain (AOG) measures whether the weak-to-strong improvement transfers beyond the source domain:
\[
    \mathrm{AOG}(S,T;m) = A_m(S,T) - A_w(S,T).
\]
Positive AOG indicates that the strong student outperforms the weak supervisor under zero-shot preference-domain shift.

\paragraph{Net Transfer Score.}
OOD improvement is desirable only if it does not come at the cost of source-domain collapse. We therefore define the in-distribution regression cost as
\[
    C_{\mathrm{ID}}(S;m) = \max\left(0, A_w(S,S) - A_m(S,S)\right),
\]
and define the Net Transfer Score (NTS) as
\[
    \mathrm{NTS}(S,T;m) = \mathrm{AOG}(S,T;m) - C_{\mathrm{ID}}(S;m).
\]
When the strong student preserves or improves source-domain performance relative to the weak supervisor, \(C_{\mathrm{ID}}(S;m)=0\), and NTS equals AOG. When target-domain gains are obtained by falling below the weak supervisor on the source domain, NTS penalizes this regression. Thus, NTS captures the central requirement of zero-shot weak-to-strong transfer: improving target-domain performance without sacrificing source-domain reliability.

%% file: content/04_method.tex
\section{Methodology}
\label{sec:method}

\begin{figure*}[t]
    \centering
    \includegraphics[width=\linewidth]{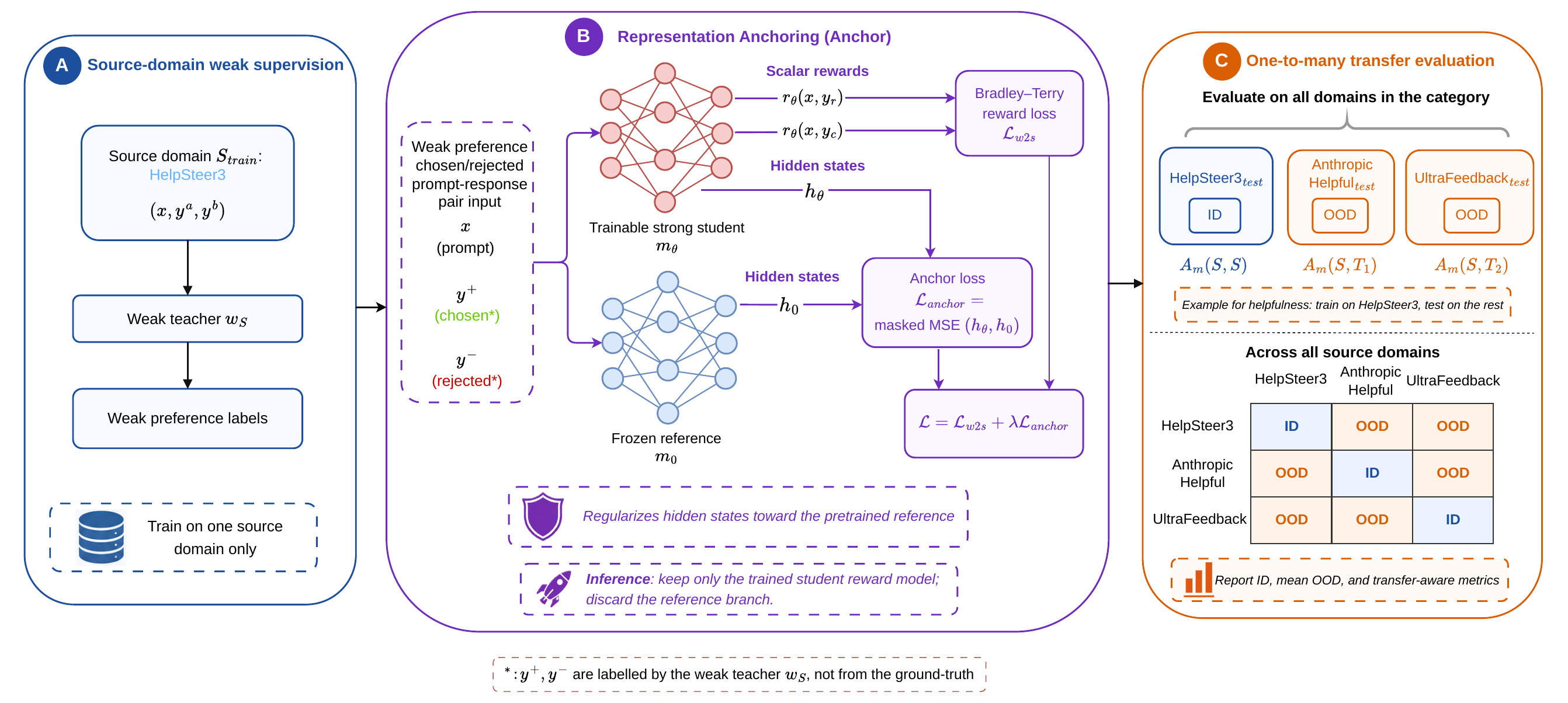}
    \caption{\textbf{Overview of \textsc{Anchor} under zero-shot preference-domain shift.}
    \textsc{Anchor} trains a strong reward model with the standard weak-to-strong preference loss \(\mathcal{L}_{\mathrm{w2s}}\), while regularizing its response-token hidden states toward a frozen pretrained reference model through \(\mathcal{L}_{\mathrm{anchor}}\). At inference, the reference model is discarded and only the learned scalar reward model is used for both in-distribution and zero-shot out-of-distribution evaluation.}
    \label{fig:method}
\end{figure*}

We now describe \textsc{Anchor}, our weak-to-strong method for zero-shot preference-domain transfer. As shown in Figure~\ref{fig:method}, \textsc{Anchor} augments standard reward-model training with a representation anchoring regularizer. The student is trained to match weak teacher preferences through the usual scalar reward modeling loss, while its response-token hidden states are simultaneously constrained to remain close to those of a frozen pretrained reference model. The final objective combines the preference loss (\(\mathcal{L}_{w2s}\)) and the anchoring loss (\(\mathcal{L}_{anchor}\)), allowing the student to learn from weak supervision while discouraging excessive deviation from pretrained representations that may support transfer.

\subsection{Weak Preference Learning Objective}
\label{sec:weak_preference_objective}

Let $\mathcal{P}_S=\{(x_i,y_i^a,y_i^b)\}_{i=1}^N$ denote a set of source-domain prompt--response pairs, where $x_i$ is a prompt and $y_i^a,y_i^b$ are two candidate responses. In the weak-to-strong setting, the strong student does not observe the ground-truth preference label. Instead, a weak teacher reward model $w_S$, trained on source domain $S$, provides the soft preference target
\[
q_i = \sigma\left(w_S(x_i,y_i^a)-w_S(x_i,y_i^b)\right),
\]
where $q_i$ is the teacher-assigned probability that $y_i^a$ is preferred to $y_i^b$.

The strong student $m_S$ predicts
\[
p_i = \sigma\left(m_S(x_i,y_i^a)-m_S(x_i,y_i^b)\right).
\]
Following the soft-label W2S objective of \citet{burns2023weakstrong}, let $z_i\sim\mathrm{Bernoulli}(q_i)$ denote the preference label induced by the weak teacher. The expected negative log-likelihood under this teacher distribution is
\[
\mathbb{E}_{z_i}\left[-\log\,p_{\theta}(z_i)\right] = -q_i\log p_i-(1-q_i)\log(1-p_i).
\]

Averaging over the training examples gives
\[
\mathcal{L}_{\mathrm{w2s}} = -\frac{1}{N}\sum_{i=1}^N \left[q_i\log p_i+(1-q_i)\log(1-p_i)\right].
\]

$\mathcal{L}_{\mathrm{w2s}}$ trains the strong student to match the weak teacher's soft preference distribution while retaining the standard scalar reward-modeling form, in which each prompt--response pair receives an independent score and pairwise preferences are derived from score differences. In \textsc{Anchor}, this objective provides the preference-learning signal and is combined with the representation-anchoring regularizer to encourage zero-shot transfer across preference domains.

\subsection{Representation Anchoring Regularizer}
\label{sec:representation_anchor_regularizer}

Our central hypothesis is that the pretrained strong model contains broadly useful representations for preference judgment, while weak-supervised fine-tuning on a single source domain may move these representations toward source-specific features. \textsc{Anchor} addresses this by adding a frozen copy of the pretrained strong model as a training-time reference. Let \(m_S\) denote the trainable strong reward model, and let \(m_{\mathrm{ref}}\) denote a frozen copy initialized from the same pretrained checkpoint before weak-supervised fine-tuning. The reference model is never updated and is discarded after training.

For a prompt--response pair \((x,y)\), we denote
\[
    H_\theta^\ell(x,y) \in \mathbb{R}^{T \times d},
    \qquad
    H_{\mathrm{ref}}^\ell(x,y) \in \mathbb{R}^{T \times d}
\]
as the hidden states at layer \(\ell\) from the student and reference models, where \(T\) is the sequence length and \(d\) is the hidden dimension. We use a response-token mask
\[
    \mathrm{m}(x,y) \in \{0,1\}^{T}
\]
to exclude prompt tokens and padding tokens from the anchoring loss.  Following that, the masked hidden-state distance at layer \(\ell\) is defined as
\[
D^\ell_{\mathrm{anchor}}(x,y)
=
\frac{\sum_{t=1}^{T}
\mathrm{m}_t(x,y)
\left\|
\Delta_t^\ell(x,y)
\right\|_2^2}{d\sum_{t=1}^{T} \mathrm{m}_t(x,y)} ,
\]

where

\[
    \Delta_t^\ell(x,y) = H_{\theta,t}^\ell(x,y) - H_{\mathrm{ref},t}^\ell(x,y).
\]

For the \(i\)-th pseudo-preference pair, the anchoring loss is computed on both candidate responses:
\[
\mathcal{L}_{\mathrm{anchor}}^\ell(i)
=
D^\ell_{\mathrm{anchor}}(x_i,y_i^a)
+
D^\ell_{\mathrm{anchor}}(x_i,y_i^b).
\]
Therefore, the full \textsc{Anchor} objective on the \(i\)-th sample is
\[
\mathcal{L}(i) = \mathcal{L}_{\mathrm{w2s}}(i)
+
\lambda
\mathcal{L}_{\mathrm{anchor}}^\ell(i),
\]
where \(\lambda\) controls the strength of representation preservation.

The two terms play complementary roles. The weak preference loss provides the task signal by training \(m_S\) to reproduce the weak teacher's soft preferences, whereas the anchoring loss regularizes the internal response representations used to support these predictions. By regularizing the student's representation space close to the pretrained reference space during source-domain fine-tuning, \textsc{Anchor} aims to preserve broadly useful pretrained features while still allowing the reward model to adapt to weak supervision.

\subsection{\textsc{Anchor} Variants}
\label{sec:anchor_variants}

We consider two variants of \textsc{Anchor} that differ in which layer representations are regularized.

\paragraph{Last-Layer \textsc{Anchor}.}
Our default variant applies the anchoring loss to the final transformer layer:
\[
\mathcal{L}_{\mathrm{Last}}(i)
=
\mathcal{L}_{\mathrm{anchor}}^{L}(i),
\]
where \(L\) is the number of transformer layers. This variant directly regularizes the representation used by the reward head and requires storing hidden states from only one layer. Because of its lower memory and computation cost, we use Last-Layer \textsc{Anchor} as the main version of our method.

\paragraph{Middle-Layer \textsc{Anchor}.}
We also evaluate a middle-layer variant, which anchors intermediate representations while leaving the final reward-facing layers more flexible. Let \(\mathcal{O}\) be a small set of normalized layer positions, for example
\[
\mathcal{O}
=
\left\{
\left\lfloor 0.5L \right\rfloor,
\left\lfloor 0.75L \right\rfloor
\right\}.
\]
The middle-layer anchoring loss is
\[
\mathcal{L}_{\mathrm{MLA}}(i)
=
\frac{1}{|\mathcal{O}|}
\sum_{\ell \in \mathcal{O}}
\mathcal{L}_{\mathrm{anchor}}^\ell(i).
\]
The corresponding training objective now becomes
\[
\mathcal{L}(i) = \mathcal{L}_{\mathrm{w2s}}(i) + \lambda \mathcal{L}_{\mathrm{MLA}}(i).
\]
Middle-Layer \textsc{Anchor} is useful both as an alternative regularizer and as a diagnostic. If preserving intermediate representations improves OOD transfer, it provides evidence consistent with a role for representation preservation in weak-to-strong transfer. In our main experiments, we use Last-Layer \textsc{Anchor} for efficiency and compare it with Middle-Layer \textsc{Anchor} in the ablation study.


%% file: content/05_exp.tex
\section{Experiment}
\label{sec:experiment}

\subsection{Experimental Setup}
\label{sec:experimental_setup}

\paragraph{Datasets.}
We evaluate on two preference categories. For \textbf{Helpful}, we use Anthropic Helpful from HH-RLHF \citep{bai2022training}, HelpSteer3-Preference \citep{wang2025helpsteer3preference}, and UltraFeedback \citep{cui2024ultrafeedback}. For \textbf{Harmless}, we use Anthropic Harmless from HH-RLHF \citep{bai2022training}, PKU-SafeRLHF \citep{ji2024pku}, and RAIL \citep{verma2025rail}. When a dataset does not provide a validation split, we reserve 10\% of its training split for validation. We then split the remaining training data into two equal parts: a gold-labeled subset \(\mathcal{D}^{\mathrm{gold}}_S\), used to train the weak teacher \(w_S\), and a weak-to-strong subset \(\mathcal{D}^{\mathrm{w2s}}_S\), which is relabeled by \(w_S\) and used to train the strong student \(m_S\).

\paragraph{Models.}
We report results on two model families. For Llama, we use Llama-3.2-1B-Instruct as the weak teacher and Llama-3.1-8B-Instruct as the strong student \citep{dubey2024llama}. For Qwen, we use Qwen3-1.7B as the weak teacher and Qwen3-8B as the strong student \citep{yang2025qwen3}. All models are trained as scalar reward models using LoRA with $\text{rank}=16$. Details of the hyperparameters can be found in Appendix~\ref{app:hyperparameter}.

\paragraph{Baselines.}
We compare \textsc{Anchor} with three W2S preference-learning baselines:
\begin{itemize}
    \item \textbf{Naive W2S} \citep{burns2023weakstrong}: trains the strong student directly on weak teacher labels using the standard weak preference loss \(\mathcal{L}_{\mathrm{w2s}}\) from Section~\ref{sec:weak_preference_objective}.
    \item \textbf{Confidence-based W2S} \citep{burns2023weakstrong}: augments naive W2S with an auxiliary confidence loss that encourages confident student predictions and reduces direct imitation of weak-teacher errors.
    \item \textbf{SEAM} \citep{li2025strongempowered}: uses the pretrained strong model to generate principle-based annotation rationales, while the weak model selects the final weak annotation.
\end{itemize}

\paragraph{Metrics and reporting.}
We use pairwise preference accuracy as the base measure and report the full accuracy table in Appendix~\ref{app:full_results}. In the main paper, we summarize performance using PGR \citep{burns2023weakstrong} and three transfer-aware metrics: WRG, AOG, and NTS, all computed from mean accuracies over three random seeds.

\subsection{Main results}

\begin{table*}[t]
\centering
\caption{\textbf{Weak-to-strong reward-model transfer under preference-domain shift.}
We report in-distribution and zero-shot out-of-distribution transfer results across helpful and harmless preference domains.}
\label{tab:main_result}
\setlength{\tabcolsep}{2.2pt}
\renewcommand{\arraystretch}{1.06}
\resizebox{\textwidth}{!}{%
\begin{tabular}{@{}l*{18}{r}@{}}
\toprule
\multicolumn{19}{@{}c}{\large\textbf{Helpful category}}\\
\midrule
\multicolumn{19}{@{}l}{\emph{Llama-3.2-1B-Instruct $\rightarrow$ Llama-3.1-8B-Instruct}}\\
& \multicolumn{6}{c}{H3} & \multicolumn{6}{c}{AH} & \multicolumn{6}{c}{UF}\\
\cmidrule(lr){2-7}\cmidrule(lr){8-13}\cmidrule(l){14-19}
Method & \multicolumn{2}{c}{In-dist.} & \multicolumn{2}{c}{AH} & \multicolumn{2}{c}{UF} & \multicolumn{2}{c}{In-dist.} & \multicolumn{2}{c}{H3} & \multicolumn{2}{c}{UF} & \multicolumn{2}{c}{In-dist.} & \multicolumn{2}{c}{H3} & \multicolumn{2}{c}{AH}\\
& PGR (\%) & WRG & AOG & NTS & AOG & NTS & PGR (\%) & WRG & AOG & NTS & AOG & NTS & PGR (\%) & WRG & AOG & NTS & AOG & NTS\\
\midrule
Naive & \underline{32.23} & \underline{1.71} & 1.46 & 1.46 & 4.8 & 4.8 & \underline{98.7} & \underline{3.13} & \textbf{4.59} & \textbf{4.59} & \underline{4.98} & \underline{4.98} & \underline{73.4} & \underline{1.74} & \underline{5.79} & \underline{5.79} & \underline{0.6} & \underline{0.6}\\
Conf.-based & -39.9 & -2.12 & 8.06 & \underline{5.94} & \underline{7.64} & \underline{5.52} & 70.3 & 2.23 & \underline{3.83} & \underline{3.83} & 4.74 & 4.74 & 41.8 & 0.99 & 3.39 & 3.39 & 0.25 & 0.25\\
SEAM & -295.1 & -15.67 & \textbf{9.95} & -5.72 & 6.25 & -9.42 & -411.4 & -13.04 & -6.01 & -19.05 & -0.46 & -13.5 & -280.6 & -6.65 & -5.68 & -12.33 & 0.17 & -6.48\\
\textsc{Anchor} & \textbf{67.2} & \textbf{3.57} & \underline{8.19} & \textbf{8.19} & \textbf{7.93} & \textbf{7.93} & \textbf{106.9} & \textbf{3.39} & 2.08 & 2.08 & \textbf{5.61} & \textbf{5.61} & \textbf{97.9} & \textbf{2.32} & \textbf{6.01} & \textbf{6.01} & \textbf{4.41} & \textbf{4.41}\\
\addlinespace[0.35em]
\hdashline
\multicolumn{19}{@{}l}{\emph{Qwen3-1.7B $\rightarrow$ Qwen3-8B}}\\
& \multicolumn{6}{c}{H3} & \multicolumn{6}{c}{AH} & \multicolumn{6}{c}{UF}\\
\cmidrule(lr){2-7}\cmidrule(lr){8-13}\cmidrule(l){14-19}
Method & \multicolumn{2}{c}{In-dist.} & \multicolumn{2}{c}{AH} & \multicolumn{2}{c}{UF} & \multicolumn{2}{c}{In-dist.} & \multicolumn{2}{c}{H3} & \multicolumn{2}{c}{UF} & \multicolumn{2}{c}{In-dist.} & \multicolumn{2}{c}{H3} & \multicolumn{2}{c}{AH}\\
& PGR (\%) & WRG & AOG & NTS & AOG & NTS & PGR (\%) & WRG & AOG & NTS & AOG & NTS & PGR (\%) & WRG & AOG & NTS & AOG & NTS\\
\midrule
Naive & -87.7 & -3.06 & \underline{-0.04} & \underline{-3.1} & 0.17 & -2.89 & 173.1 & 0.45 & 5.57 & 5.57 & 5.15 & \underline{5.15} & -89.1 & -0.41 & \textbf{3.61} & \underline{3.2} & 1.37 & 0.96\\
Conf.-based & \underline{-65.9} & \underline{-2.3} & -1.58 & -3.88 & 4.4 & \underline{2.1} & \underline{661.5} & \underline{1.72} & \underline{6.78} & \underline{6.78} & 4.92 & 4.92 & \underline{-39.1} & \underline{-0.18} & 1.42 & 1.24 & \underline{3.26} & 3.08\\
SEAM & -332.1 & -11.59 & \textbf{8.2} & -3.39 & \textbf{5.79} & -5.8 & $-1.3k$ & -3.42 & 0.55 & -2.87 & \underline{5.49} & 2.07 & -517.4 & -2.38 & -1.85 & -4.23 & \textbf{5.88} & \textbf{3.5}\\
\textsc{Anchor} & \textbf{9.5} & \textbf{0.33} & -2.61 & \textbf{-2.61} & \underline{4.69} & \textbf{4.69} & \textbf{$\bm{1k}$} & \textbf{2.75} & \textbf{8.96} & \textbf{8.96} & \textbf{5.73} & \textbf{5.73} & \textbf{189.1} & \textbf{0.87} & \underline{3.28} & \textbf{3.28} & 3.13 & \underline{3.13} \\
\midrule
\multicolumn{19}{@{}c}{\large\textbf{Harmless category}}\\
\midrule
\multicolumn{19}{@{}l}{\emph{Llama-3.2-1B-Instruct $\rightarrow$ Llama-3.1-8B-Instruct}}\\
& \multicolumn{6}{c}{AHar} & \multicolumn{6}{c}{PKU} & \multicolumn{6}{c}{RAIL}\\
\cmidrule(lr){2-7}\cmidrule(lr){8-13}\cmidrule(l){14-19}
Method & \multicolumn{2}{c}{In-dist.} & \multicolumn{2}{c}{PKU} & \multicolumn{2}{c}{RAIL} & \multicolumn{2}{c}{In-dist.} & \multicolumn{2}{c}{AHar} & \multicolumn{2}{c}{RAIL} & \multicolumn{2}{c}{In-dist.} & \multicolumn{2}{c}{AHar} & \multicolumn{2}{c}{PKU}\\
& PGR (\%) & WRG & AOG & NTS & AOG & NTS & PGR (\%) & WRG & AOG & NTS & AOG & NTS & PGR (\%) & WRG & AOG & NTS & AOG & NTS\\
\midrule
Naive & 122.6 & 3.09 & 1.97 & 1.97 & \underline{0} & \underline{0} & 146.1 & 4.91 & \underline{2.65} & \underline{2.65} & 2.26 & 2.26 & \underline{55.7} & \underline{1.03} & \underline{2.44} & \underline{2.44} & \underline{2.04} & \underline{2.04}\\
Conf.-based & \underline{129.4} & \underline{3.26} & \underline{3.11} & \textbf{3.11} & -0.31 & -0.31 & 151.2 & 5.08 & 2.61 & 2.61 & 2.47 & 2.47 & -72.4 & -1.34 & 0.79 & -0.55 & 0.32 & -1.02\\
SEAM & -264.3 & -6.66 & \textbf{4.5} & -2.16 & -0.51 & -7.17 & \textbf{236.3} & \textbf{7.94} & 1.64 & 1.64 & \textbf{6.27} & \textbf{6.27} & -277.8 & -5.14 & -4.74 & -9.88 & -0.25 & -5.39\\
\textsc{Anchor} & \textbf{134.5} & \textbf{3.39} & 2.38 & \underline{2.38} & \textbf{0.31} & \textbf{0.31} & \underline{163.4} & \underline{5.49} & \textbf{3.43} & \textbf{3.43} & \underline{4.62} & \underline{4.62} & \textbf{77.8} & \textbf{1.44} & \textbf{2.62} & \textbf{2.62} & \textbf{2.37} & \textbf{2.37}\\
\addlinespace[0.35em]
\hdashline
\multicolumn{19}{@{}l}{\emph{Qwen3-1.7B $\rightarrow$ Qwen3-8B}}\\
& \multicolumn{6}{c}{AHar} & \multicolumn{6}{c}{PKU} & \multicolumn{6}{c}{RAIL}\\
\cmidrule(lr){2-7}\cmidrule(lr){8-13}\cmidrule(l){14-19}
Method & \multicolumn{2}{c}{In-dist.} & \multicolumn{2}{c}{PKU} & \multicolumn{2}{c}{RAIL} & \multicolumn{2}{c}{In-dist.} & \multicolumn{2}{c}{AHar} & \multicolumn{2}{c}{RAIL} & \multicolumn{2}{c}{In-dist.} & \multicolumn{2}{c}{AHar} & \multicolumn{2}{c}{PKU}\\
& PGR (\%) & WRG & AOG & NTS & AOG & NTS & PGR (\%) & WRG & AOG & NTS & AOG & NTS & PGR (\%) & WRG & AOG & NTS & AOG & NTS\\
\midrule
Naive & \textbf{160.9} & \textbf{1.48} & \underline{2.05} & \textbf{2.05} & \underline{0.52} & \underline{0.52} & $1k$ & 0.82 & \underline{1.35} & \underline{1.35} & \textbf{1.85} & \textbf{1.85} & \underline{-19.1} & \underline{-0.51} & \underline{1.53} & 1.02 & 1.72 & 1.21\\
Conf.-based & 5.4 & 0.05 & -1.72 & -1.72 & -0.51 & -0.51 & $1.5k$ & 1.23 & -0.39 & -0.39 & \underline{1.54} & \underline{1.54} & -30.7 & -0.82 & \textbf{1.88} & \underline{1.06} & \textbf{3.93} & \textbf{3.11}\\
SEAM & $-1.6$k & -14.71 & \textbf{5.4} & -9.31 & -1.54 & -16.25 & \textbf{$\bm{4.2k}$} & \textbf{3.36} & -8.49 & -8.49 & -2.67 & -2.67 & -385 & -10.28 & -10.79 & -21.07 & 1.07 & -9.21\\
\textsc{Anchor} & \textbf{160.9} & \textbf{1.48} & 0.66 & \underline{0.66} & \textbf{1.65} & \textbf{1.65} & \underline{$2.7k$} & \underline{2.21} & \textbf{1.43} & \textbf{1.43} & \textbf{1.85} & \textbf{1.85} & \textbf{0} & \textbf{0} & 1.4 & \textbf{1.4} & \underline{2.46} & \underline{2.46} \\
\bottomrule
\end{tabular}%
}
\vspace{1mm}
\\
\footnotesize{H3 = HelpSteer3, AH = Anthropic Helpful, UF = UltraFeedback, AHar = Anthropic Harmless, In-dist. = In-distribution metrics. The best value in each column and model block is shown in bold, and the second-best value is underlined. PGR is reported for comparability with prior W2S work. Because it is a ratio-based metric, it can take very large negative values when a method performs far below the weak supervisor or when the weak-to-strong gap is small.
}
\end{table*}

Table~\ref{tab:main_result} shows that in-distribution weak-to-strong gains do not always imply robust transfer to unseen preference domains. Naive and confidence-based training exhibit a similar pattern: they can achieve strong in-distribution performance, but their OOD transfer remains unstable. For example, in the harmless category with the Qwen model family, when training on Anthropic Harmless, Naive obtains tied-best in-distribution WRG score of $1.48$, but transfers weakly to RAIL with only $0.52$ in AOG. Similarly, when training on PKU-SafeRLHF, confidence-based filtering achieves a high in-distribution PGR of $1537.5\%$, but gives negative transfer to Anthropic Harmless with $-0.39$ in NTS. These results suggest that improving or filtering weak source-domain supervision alone does not reliably produce transferable reward models.

SEAM shows a sharp preservation transfer trade-off: it can achieve competitive OOD AOG, but often at the cost of severe in-distribution degradation. In the helpful Llama setting, training on HelpSteer3 gives SEAM strong OOD AOG on Anthropic Helpful and UltraFeedback ($9.95$ and $6.25$), yet its ID WRG drops to $-15.67$; the same pattern appears for Qwen trained on Anthropic Helpful, with $5.49$ OOD AOG on UltraFeedback but $-1315.4\%$ ID PGR. In contrast, \textsc{Anchor} better preserves source-domain performance while improving transfer, raising AOG from 0.60 to 4.41 on Anthropic Helpful and NTS from 5.79 to 6.01 on HelpSteer3 when trained on UltraFeedback. Across Table \ref{tab:main_result}, \textsc{Anchor} is best or tied in 41 of 60 WRG/AOG/NTS comparisons, supporting its stronger overall balance across ID and OOD evaluation rather than uniform dominance in every setting. 

Nevertheless, \textsc{Anchor} does not guarantee positive transfer for every source-target pair. In particular, Qwen trained on HelpSteer3 retains a positive ID gain ($\mathrm{WRG}=0.33$) and transfers positively to UltraFeedback ($\mathrm{NTS}=4.69$), but remains below the weak teacher on Anthropic Helpful ($\mathrm{NTS}=-2.61$). This failure persists across anchoring strengths and layer choices, suggesting that representation preservation alone cannot eliminate all source-target preference mismatch.

A comparison with the strong-supervised ceiling further contextualizes these remaining failures. The mean ID/OOD ceiling gaps are 1.01/6.37 for Naive, 1.56/6.27 for confidence-based W2S, 8.73/8.05 for SEAM, and only $-0.11/5.45$ for \textsc{Anchor}. Moreover, the additional OOD gap attributable to weak supervision is only $0.23$ points for \textsc{Anchor}, compared with $1.15$, $1.05$, and $2.83$ for the three baselines. This suggests that much of the remaining OOD deficit reflects source-target transfer difficulty that persists even under gold supervision.

\begin{table}[t]
\centering
\caption{Comparison with preservation and optimization controls on Llama/HelpSteer3. Full hyperparameter sweeps are reported in the appendix.}
\label{tab:preservation_controls}
\small
\resizebox{\linewidth}{!}{%
\begin{tabular}{lccccc}
\toprule
Method & WRG & \multicolumn{2}{c}{AH} & \multicolumn{2}{c}{UF} \\
& & AOG & NTS & AOG & NTS \\
\midrule
Naive & 1.71 & 1.46 & 1.46 & 4.80 & 4.80 \\
WI ($\alpha=0.75$) & 1.38 & 4.12 & 4.12 & 5.32 & 5.32 \\
Early stopping & 2.47 & 3.80 & 3.80 & 6.26 & 6.26 \\
LoRA ($r=8$) & 1.49 & 2.86 & 2.86 & 6.14 & 6.14 \\
L2-SP & 2.69 & 4.16 & 4.16 & \textbf{9.03} & \textbf{9.03} \\
\textsc{Anchor} & \textbf{3.57} & \textbf{8.19} & \textbf{8.19} & 7.93 & 7.93 \\
\bottomrule
\end{tabular}
}
\end{table}

\paragraph{Comparison with preservation controls.}
To test whether \textsc{Anchor}'s gains can be reproduced by simply restricting fine-tuning, we compare against weight interpolation (WI), source-validation early stopping, reduced-rank LoRA, and L2-SP parameter-space regularization in the same Llama/HelpSteer3 setting (Table~\ref{tab:preservation_controls}). The simpler controls provide partial improvements over Naive W2S, but none matches \textsc{Anchor}'s overall ID--OOD balance. In particular, early stopping and rank-$8$ LoRA reach UltraFeedback AOGs of $6.26$ and $6.14$, respectively, but remain substantially below \textsc{Anchor} on Anthropic Helpful. Parameter-space preservation is more competitive: L2-SP with $\lambda=10^{-4}$ reaches $9.03$ AOG on UltraFeedback, exceeding \textsc{Anchor}'s $7.93$, but obtains lower WRG ($2.69$ vs.\ $3.57$) and substantially weaker transfer to Anthropic Helpful ($4.16$ vs.\ $8.19$ AOG). Together, these results suggest that preserving pretrained information is beneficial, while directly constraining hidden representations provides a stronger overall preservation transfer balance than reducing optimization, limiting adaptation capacity, or constraining parameter displacement alone. Full WI and L2-SP sweeps are reported in Appendix~\ref{sec:l2sp}.

%% file: content/06_analysis.tex
\section{Ablation Studies}
\label{sec:ablation_studies}

We conduct ablation studies to better understand the role of representation anchoring in \textsc{Anchor}. Unless otherwise stated, all ablations are conducted with the Llama model family using HelpSteer3 as the source training domain. Specifically, we analyze the effect of the anchoring strength and the choice of anchored layer.

\subsection{Effect of Anchoring Strength}
\label{sec:ablation_lambda}

\begin{figure*}[t]
    \centering
    \includegraphics[width=0.8\linewidth]{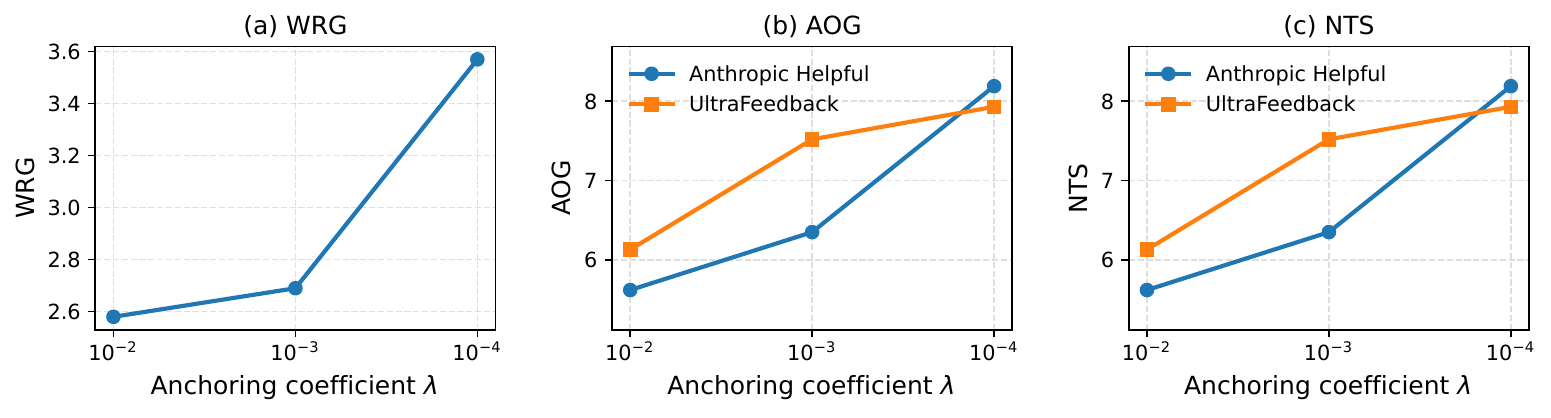}
    \caption{Effect of anchoring strength \(\lambda\) on in-distribution and out-of-distribution transfer performance. Larger x-axis values correspond to smaller anchoring coefficients, i.e., weaker representation anchoring.}
    \label{fig:lambda_ablation}
    \vspace{-5mm}
\end{figure*}

We first study the sensitivity of \textsc{Anchor} to the anchoring coefficient \(\lambda\). We vary \(\lambda \in \{10^{-2}, 10^{-3}, 10^{-4}\}\), which controls the strength of the representation-anchoring loss relative to the weak preference objective. A larger \(\lambda\) creates a stronger constraint on the student representations, which may better preserve pretrained knowledge but can also weaken adaptation to the source preference signal.

Figure~\ref{fig:lambda_ablation} shows that decreasing the anchoring coefficient consistently improves performance across WRG, AOG, and NTS. The consistent trend suggests that overly strong anchoring can constrain adaptation, whereas a weaker anchoring penalty better balances weak preference learning and representation preservation.

\subsection{Effect of Anchored Layer}
\label{sec:ablation_layer}

We further study the choice of anchoring layer. Final-layer states are closer to the reward head and may be more strongly shaped by source-domain preference learning, whereas middle-layer states may preserve more general transferable features while leaving higher layers free to adapt.

\begin{table}[t]
\centering
\caption{Effect of anchored layer with \(\lambda=10^{-4}\). Bold indicates the best value in each column.}
\label{tab:mla_anchor_1e4}
\resizebox{\linewidth}{!}{%
\begin{tabular}{lccccc}
\toprule
\multirow{2}{*}{Method} 
& \multirow{2}{*}{WRG} 
& \multicolumn{2}{c}{Anthropic Helpful} 
& \multicolumn{2}{c}{UltraFeedback} \\
\cmidrule(lr){3-4} \cmidrule(lr){5-6}
& & AOG & NTS & AOG & NTS \\
\midrule
Middle-layer
& 2.14 & 4.42 & 4.42 & \textbf{8.22} & \textbf{8.22} \\
\textsc{Anchor}
& \textbf{3.57} & \textbf{8.19} & \textbf{8.19} & 7.93 & 7.93 \\
\bottomrule
\end{tabular}%
}
\end{table}

Table~\ref{tab:mla_anchor_1e4} shows that \textsc{Anchor} provides a better overall balance across in-distribution and transfer settings. While middle-layer anchoring is slightly stronger on UltraFeedback, \textsc{Anchor} performs better in-distribution and transfers more strongly to Anthropic Helpful, suggesting that the chosen anchored layer better preserves transferable representations without overly restricting adaptation.

%% file: content/07_related_work.tex
\section{Related Work}

\paragraph{Weak-to-strong generalization and scalable oversight.}
Recent work has applied weak-to-strong (W2S) generalization to alignment and preference learning, showing that weak aligned models or weak supervisors can improve stronger models through preference optimization and related alignment objectives \citep{zhu2025weak,lyu2025macpo}. At the same time, these studies also expose limitations of weak supervision, including overfitting to weak signals or inheriting supervisor errors \citep{shi2025mitigate,yang2024superficial}. Beyond preference supervision, W2S has also been studied in reasoning, mathematical problem solving, weak-guided search, and agentic decision-making, suggesting that weak supervision is a broader paradigm for eliciting strong-model capabilities \citep{yang2024weak,zhou2024weakstrongsearch,nie2025weakforstrong}.

\paragraph{Preference learning and reward modeling.}
Learning scalar reward models from comparisons is a core component of RLHF-style alignment \citep{christiano2017deep,ouyang2022training,bai2022training}. Most reward-modeling objectives use a Bradley--Terry likelihood over response pairs \citep{bradley1952rank}, and recent benchmarks further treat reward models as standalone evaluators across chat, reasoning, and safety settings \citep{lambert2025rewardbench}. In parallel, direct preference-optimization methods bypass explicit reward-model training for policy optimization \citep{rafailov2023direct,gheshlaghi2024general,hong2024orpo,meng2024simpo}. Joint evaluator models instead compare candidate responses together, including pairwise rankers, comparative LLM judges, and evaluator LMs with pairwise-ranking modes \citep{jiang2023llmblender,liusie2024llm,li2024generative,kim2024prometheus}; however, pair-dependent scoring is less compatible with settings that require independent candidate scores.

\paragraph{Representation-preserving fine-tuning.}
Fine-tuning can degrade pretrained features and OOD robustness \citep{kumar2022finetuning,li2024dual,zang2024overcoming}, related to catastrophic forgetting in continual learning \citep{li2024revisiting,zheng2025spurious}. Prior approaches preserve knowledge through distillation \citep{hinton2015distilling}, continual-learning regularization \citep{kirkpatrick2017overcoming,li2017learning,wu2024meta}, or parameter-space constraints \citep{lee2020mixout}. Unlike continual learning, which typically preserves performance on previously observed tasks, our setting evaluates zero-shot transfer from a single source preference domain to unseen domains. \textsc{Anchor} adapts representation preservation to W2S reward modeling by anchoring response-token hidden states to a frozen pretrained reference, with no inference-time modification.

%% file: content/08_conclusion.tex
\section{Conclusion}


We study weak-to-strong reward modeling under zero-shot preference-domain shift and introduce an evaluation protocol that trains on one preference domain while testing both in-distribution and on unseen target domains. Using this protocol, we show that in-distribution W2S gains do not necessarily indicate robust alignment: methods that improve source-domain performance may still transfer poorly across domains. To address this gap, we propose \textsc{Anchor}, a representation-anchoring method regularizing the strong student toward a frozen pretrained reference during weak-supervised training. Across preference domains and model families, \textsc{Anchor} better preserves in-distribution performance while improving out-of-distribution transfer. Our findings suggest that robust W2S alignment should be evaluated not only by whether strong models learn from weak supervision, but also by whether the learned reward function remains reliable beyond the training domain.

%% file: content/09_limitation.tex
\section*{Limitations}

Our study focuses primarily on offline pairwise reward-model accuracy across a fixed suite of helpful and harmless preference datasets. While this controlled setting allows us to systematically analyze preference-domain transfer, offline metrics may not fully capture downstream deployment behavior. We provide initial downstream validation through best-of-4 reranking (Appendix~\ref{sec:bon}), where \textsc{Anchor} improves candidate selection across both ID and OOD settings. However, broader downstream evaluation through online RLHF policy optimization, DPO-style policy training, and open-ended generation is left for future work.

Our cross-dataset evaluation also represents a compound preference-domain shift: the datasets may differ simultaneously in prompt distributions, response styles and sources, annotation protocols, and preference policies. Consequently, our experiments do not disentangle the contribution of each factor to OOD failure, nor establish representation drift as its sole cause. Isolating these sources of shift through more controlled evaluations is left for future work.

Furthermore, we note a distinct trade-off regarding computational efficiency. In a representative Llama/UltraFeedback run, \textsc{Anchor} increases training time and GPU-hours by approximately 14\% relative to Naive W2S (4.11 vs. 3.62 minutes; 0.069 vs. 0.060 GPU-hours), while measured peak VRAM remains effectively unchanged (38.66 vs. 38.71 GB/GPU). This overhead comes from one additional frozen-reference forward pass, with no reference-model backward pass and no inference-time overhead. Finally, our empirical validation is limited to specific helpful/harmless domains and specific model scales; future work is required to evaluate \textsc{Anchor} across broader preference axes, more diverse task domains, and larger model families.

%% file: content/10_ack.tex
\section*{Acknowledgements}

This research/project is supported by the National Research Foundation, Singapore under
its AI Singapore Programme (AISG Award No: AISG3-PhD-2025-08-058).

%% file: content/appendix.tex
\input{content/appendix_materials/app_a}

\input{content/appendix_materials/app_b}

\input{content/appendix_materials/app_c}

\input{content/appendix_materials/app_d}

\input{content/appendix_materials/app_e}

\input{content/appendix_materials/app_f}

\section{Downstream Best-of-4 Reranking Evaluation}
\label{sec:bon}

To examine whether improvements in pairwise reward-model accuracy translate to downstream candidate selection, we conduct a best-of-4 reranking evaluation on UltraFeedback using the Llama reward models. We evaluate models trained on HelpSteer3, Anthropic Helpful, and UltraFeedback. The first two settings therefore measure zero-shot OOD transfer to UltraFeedback, while training on UltraFeedback provides the corresponding ID setting. We compare Naive W2S, confidence-based W2S, and \textsc{Anchor}, and report both best-of-4 top-1 accuracy and pairwise accuracy in Table~\ref{tab:bon_results}.

\begin{table}[t]
\centering
\caption{Best-of-4 reranking on UltraFeedback using Llama reward models. HelpSteer3 and Anthropic Helpful are OOD source domains, while UltraFeedback is the ID source. Higher is better.}
\label{tab:bon_results}
\small
\begin{tabular}{llcc}
\toprule
Source & Method & Bo4 Top-1 & Pairwise \\
\midrule
\multirow{3}{*}{HelpSteer3}
& Naive       & 45.17 & 69.58 \\
& Conf.-based & 47.09 & 70.53 \\
& \textsc{Anchor} & \textbf{48.86} & \textbf{72.14} \\
\midrule
\multirow{3}{*}{Anthropic Helpful}
& Naive       & 44.18 & 70.31 \\
& Conf.-based & 44.21 & 70.51 \\
& \textsc{Anchor} & \textbf{45.24} & \textbf{70.68} \\
\midrule
\multirow{3}{*}{UltraFeedback}
& Naive       & 53.55 & 76.37 \\
& Conf.-based & 52.84 & 75.45 \\
& \textsc{Anchor} & \textbf{55.47} & \textbf{77.44} \\
\bottomrule
\end{tabular}
\end{table}

\textsc{Anchor} achieves the highest best-of-4 top-1 accuracy in all three settings, including both OOD transfers. Averaged across source domains, it improves top-1 accuracy by $2.22$ points over Naive W2S and $1.81$ points over confidence-based W2S. The same ordering is observed for pairwise accuracy. These results provide initial evidence that the transfer improvements observed in offline reward-model evaluation can translate into better downstream candidate selection. Broader evaluation through RLHF, DPO-style policy training, and open-ended generation is left for future work.




%% file: content/appendix_materials/app_a.tex
\section{Hyperparameters and Implementation Details}
\label{app:hyperparameter}

We use LoRA fine-tuning for all weak-to-strong methods, including both the weak teacher and the strong student models. Unless otherwise specified, we use the final checkpoint after three epochs for evaluation and apply the same training configuration across methods:
\begin{itemize}
    \item LoRA rank: 16
    \item Learning rate: $2 \times 10^{-4}$
    \item Number of epochs: 3
    \item Anchoring coefficient for \textsc{Anchor}: $\lambda = 10^{-4}$
\end{itemize}

For \textsc{Anchor}, we select $\lambda = 10^{-4}$ from a preliminary sweep over $\lambda \in \{10^{-2}, 10^{-3}, 10^{-4}\}$ on the HelpSteer3 source-domain setting, where it yields the strongest in-domain validation performance. We then fix this value across all datasets and model families to avoid per-setting hyperparameter tuning.

All experiments were conducted on NVIDIA RTX A6000 GPUs. Reproducing the reported results requires approximately 2400 GPU-hours in total, including runs across datasets, random seeds, and baselines. To quantify the additional cost of representation anchoring, we report training efficiency for a representative Llama/UltraFeedback run in Table~\ref{tab:training_time_measurement}.

\begin{table}[h]
\centering
\caption{Training overhead on a representative Llama/UltraFeedback run. Relative time and VRAM are normalized to Naive W2S. \textsc{Anchor} adds approximately 14\% training time and GPU-hours, while measured peak VRAM remains effectively unchanged.}
\label{tab:training_time_measurement}
\small
\resizebox{\linewidth}{!}{%
\begin{tabular}{lrrrrr}
\toprule
Method & Time (min) & VRAM (GB) & GPU-hours & Relative & Relative \\
 &  & & & Time & VRAM \\
\midrule
Naive W2S   & 3.62 & 38.71 & 0.060 & 1.00$\times$ & 1.00$\times$ \\
Conf.-based & 3.48 & 38.71 & 0.058 & 0.96$\times$ & 1.00$\times$ \\
\textsc{Anchor} & 4.11 & 38.66 & 0.069 & 1.14$\times$ & 1.00$\times$ \\
\bottomrule
\end{tabular}%
}
\end{table}

%% file: content/appendix_materials/app_b.tex
\section{Data Processing Pipeline}

Most preference-learning datasets provide a prompt, a chosen response, a rejected response, and the corresponding chosen and rejected scores. We first remove samples where the chosen and rejected scores are equal, since these tied examples do not provide a clear preference signal. We then concatenate each prompt with its chosen and rejected responses to construct a chosen conversation and a rejected conversation, respectively. These paired conversations are used as the input examples for reward model training and evaluation.

%% file: content/appendix_materials/app_c.tex
\section{Dataset Statistics and Preference-Domain Shift Diagnostics}

\subsection{Dataset split statistics}

Table~\ref{tab:dataset_statistics} summarizes the dataset statistics used in our experiments, including the train, validation, test, weak-teacher training, and weak-to-strong training splits for each preference dataset.

\begin{table}[ht]
\centering
\caption{Dataset statistics used in our weak-to-strong reward-modeling experiments. The training split is divided equally into weak-teacher training data and weak-to-strong training data.}
\label{tab:dataset_statistics}
\small
\resizebox{\linewidth}{!}{%
\begin{tabular}{llrrrrr}
\toprule
Dataset & Category & Train & Validation & Test & Weak-teacher train & W2S train \\
\midrule
Anthropic Helpful & Helpful & 115,396 & 11,540 & 2,332 & 57,698 & 57,698 \\
HelpSteer3 & Helpful & 17,708 & 1,771 & 915 & 8,854 & 8,854 \\
UltraFeedback & Helpful & 53,748 & 5,375 & 1,728 & 26,874 & 26,874 \\
\midrule
Anthropic Harmless & Harmless & 42,254 & 4,226 & 2,298 & 21,127 & 21,127 \\
PKU-SafeRLHF & Harmless & 26,874 & 2,688 & 1,222 & 13,437 & 13,437 \\
RAIL & Harmless & 7,862 & 887 & 973 & 3,931 & 3,931 \\
\bottomrule
\end{tabular}%
}
\end{table}

\subsection{Cross-dataset shift diagnostics}

To characterize the distribution differences underlying our cross-dataset evaluation, Table~\ref{tab:dataset_shift} reports simple statistics of the prompt and preferred-response distributions. Token lengths are measured using the Llama-3 tokenizer. Even within the same broad preference category, these datasets differ substantially in prompt and response length. For example, HelpSteer3 contains considerably longer prompts and preferred responses than Anthropic Helpful, while the harmlessness datasets also differ in both response length and the frequency with which the preferred response is longer than the rejected response.

\begin{table}[ht]
\centering
\caption{Cross-dataset preference-domain diagnostics. Prompt and chosen-response lengths are mean token counts measured with the Llama-3 tokenizer. ``Chosen longer'' denotes the fraction of preference pairs in which the preferred response is longer than the rejected response.}
\label{tab:dataset_shift}
\small
\resizebox{\linewidth}{!}{%
\begin{tabular}{lrrr}
\toprule
Dataset & Prompt len. & Chosen len. & Chosen longer \\
\midrule
Anthropic Helpful  & 162 &  56 & 59\% \\
HelpSteer3         & 565 & 389 & 56\% \\
UltraFeedback      & 182 & 275 & 56\% \\
\midrule
Anthropic Harmless & 149 &  38 & 43\% \\
PKU-SafeRLHF       &  61 &  82 & 38\% \\
RAIL               & 137 &  47 & 34\% \\
\bottomrule
\end{tabular}
}
\end{table}

The datasets additionally differ in response provenance and annotation policy. Anthropic HH-RLHF is based on binary human preference comparisons. Meanwhile, HelpSteer3 uses graded preferences collected from multiple annotators and UltraFeedback derives preferences from multidimensional GPT-4-based evaluations. In the harmlessness category, PKU-SafeRLHF focuses on preferences between safer and less-safe responses while RAIL uses multidimensional responsible-AI criteria. Consequently, our cross-dataset setting jointly varies prompt distributions, response styles and sources, and preference-labeling policies.

These diagnostics support treating the evaluation as a compound preference-domain shift rather than an ordinary held-out dataset split. However, because these factors vary jointly, they do not identify any single factor, including representation drift, response-distribution shift, or annotation-policy differences, as the sole cause of the observed OOD degradation.

%% file: content/appendix_materials/app_d.tex
\section{Full Accuracy Results and Strong-Ceiling Analysis}
\label{app:full_results}

\subsection{Full Accuracy Results}

Table~\ref{tab:acc_std_full} reports the full accuracy results with standard deviations across different reward-modeling methods, datasets, and model families.

\begin{table*}[t]
\centering
\scriptsize
\setlength{\tabcolsep}{3.2pt}
\renewcommand{\arraystretch}{1.12}
\caption{Accuracy results across helpful and harmless preference-domain shifts. Each three-column block indicates the source training dataset, and each subcolumn indicates the evaluation dataset. Standard deviations over three seeds are shown as subscripts.}
\label{tab:acc_std_full}
\resizebox{\linewidth}{!}{%
\begin{tabular}{@{}lccccccccc@{}}
\toprule
\multicolumn{10}{c}{\textbf{Helpful}} \\
\midrule
\multicolumn{10}{l}{\emph{Llama-3.2-1B-Instruct $\rightarrow$ Llama-3.1-8B-Instruct}} \\
& \multicolumn{3}{c}{\textbf{Src: HelpSteer3}} 
& \multicolumn{3}{c}{\textbf{Src: Anthropic Helpful}} 
& \multicolumn{3}{c}{\textbf{Src: UltraFeedback}} \\
\cmidrule(lr){2-4} \cmidrule(lr){5-7} \cmidrule(lr){8-10}
\textbf{Method} 
& HelpSteer3 & AHelpful & UltraFB
& AHelpful & HelpSteer3 & UltraFB
& UltraFB & HelpSteer3 & AHelpful \\
\midrule
Weak teacher     
& \accstd{74.03}{0.77} & \accstd{52.96}{0.59} & \accstd{63.89}{0.23}
& \accstd{69.47}{0.51} & \accstd{64.26}{0.23} & \accstd{65.51}{0.72}
& \accstd{74.94}{0.49} & \accstd{64.48}{0.72} & \accstd{62.14}{0.72} \\
Strong ceiling   
& \accstd{79.34}{0.37} & \accstd{60.98}{0.77} & \accstd{73.50}{0.40}
& \accstd{72.64}{0.59} & \accstd{67.76}{0.72} & \accstd{71.47}{0.66}
& \accstd{77.31}{0.26} & \accstd{71.26}{0.64} & \accstd{64.67}{0.77} \\
Naive            
& \accstd{75.74}{0.72} & \accstd{54.42}{0.40} & \accstd{68.69}{0.71}
& \accstd{72.60}{0.69} & \accstd{68.85}{0.72} & \accstd{70.49}{0.77}
& \accstd{76.68}{0.28} & \accstd{70.27}{0.31} & \accstd{62.74}{0.77} \\
Confidence-based 
& \accstd{71.91}{0.72} & \accstd{61.02}{0.24} & \accstd{71.53}{0.45}
& \accstd{71.70}{0.19} & \accstd{68.09}{0.77} & \accstd{70.25}{0.72}
& \accstd{75.93}{0.21} & \accstd{67.87}{0.72} & \accstd{62.39}{0.72} \\
SEAM             
& \accstd{58.36}{0.77} & \accstd{62.91}{0.70} & \accstd{70.14}{0.50}
& \accstd{56.43}{0.77} & \accstd{58.25}{0.77} & \accstd{65.05}{0.59}
& \accstd{68.29}{0.76} & \accstd{58.80}{0.72} & \accstd{62.31}{0.75} \\
\textsc{Anchor}  
& \accstd{77.60}{0.38} & \accstd{61.15}{0.52} & \accstd{71.82}{0.43}
& \accstd{72.86}{0.47} & \accstd{66.34}{0.77} & \accstd{71.12}{0.72}
& \accstd{77.84}{0.29} & \accstd{70.49}{0.58} & \accstd{66.55}{0.72} \\
\hdashline
\multicolumn{10}{l}{\emph{Qwen3-1.7B $\rightarrow$ Qwen3-8B}} \\
& \multicolumn{3}{c}{\textbf{Src: HelpSteer3}} 
& \multicolumn{3}{c}{\textbf{Src: Anthropic Helpful}} 
& \multicolumn{3}{c}{\textbf{Src: UltraFeedback}} \\
\cmidrule(lr){2-4} \cmidrule(lr){5-7} \cmidrule(lr){8-10}
\textbf{Method} 
& HelpSteer3 & AHelpful & UltraFB
& AHelpful & HelpSteer3 & UltraFB
& UltraFB & HelpSteer3 & AHelpful \\
\midrule
Weak teacher     
& \accstd{75.74}{0.72} & \accstd{59.43}{0.32} & \accstd{67.42}{0.41}
& \accstd{71.05}{0.65} & \accstd{61.64}{0.16} & \accstd{68.58}{0.72}
& \accstd{76.97}{0.26} & \accstd{67.21}{0.50} & \accstd{62.09}{0.72} \\
Strong ceiling   
& \accstd{79.23}{0.72} & \accstd{58.40}{0.72} & \accstd{71.88}{0.25}
& \accstd{71.31}{0.72} & \accstd{65.36}{0.25} & \accstd{75.23}{0.53}
& \accstd{77.43}{0.62} & \accstd{69.73}{0.77} & \accstd{62.99}{0.34} \\
Naive            
& \accstd{72.68}{0.72} & \accstd{59.39}{0.27} & \accstd{67.59}{0.72}
& \accstd{71.50}{0.16} & \accstd{67.21}{0.35} & \accstd{73.73}{0.77}
& \accstd{76.56}{0.15} & \accstd{70.82}{0.50} & \accstd{63.46}{0.18} \\
Confidence-based 
& \accstd{73.44}{0.49} & \accstd{57.85}{0.35} & \accstd{71.82}{0.77}
& \accstd{72.77}{0.72} & \accstd{68.42}{0.49} & \accstd{73.50}{0.55}
& \accstd{76.79}{0.33} & \accstd{68.63}{0.72} & \accstd{65.35}{0.20} \\
SEAM             
& \accstd{64.15}{0.22} & \accstd{67.63}{0.77} & \accstd{73.21}{0.49}
& \accstd{67.63}{0.72} & \accstd{62.19}{0.72} & \accstd{74.07}{0.72}
& \accstd{74.59}{0.19} & \accstd{65.36}{0.20} & \accstd{67.97}{0.33} \\
\textsc{Anchor}  
& \accstd{76.07}{0.72} & \accstd{56.82}{0.77} & \accstd{72.11}{0.77}
& \accstd{73.80}{0.76} & \accstd{70.60}{0.72} & \accstd{74.31}{0.56}
& \accstd{77.84}{0.72} & \accstd{69.73}{0.72} & \accstd{65.22}{0.33} \\
\midrule

\multicolumn{10}{c}{\textbf{Harmless}} \\
\midrule
\multicolumn{10}{l}{\emph{Llama-3.2-1B-Instruct $\rightarrow$ Llama-3.1-8B-Instruct}} \\
& \multicolumn{3}{c}{\textbf{Src: Anthropic Harmless}} 
& \multicolumn{3}{c}{\textbf{Src: PKU}} 
& \multicolumn{3}{c}{\textbf{Src: RAIL}} \\
\cmidrule(lr){2-4} \cmidrule(lr){5-7} \cmidrule(lr){8-10}
\textbf{Method} 
& AHarmless & PKU & RAIL
& PKU & AHarmless & RAIL
& RAIL & AHarmless & PKU \\
\midrule
Weak teacher     
& \accstd{70.02}{0.77} & \accstd{90.75}{0.72} & \accstd{88.49}{0.37}
& \accstd{85.92}{0.64} & \accstd{61.58}{0.46} & \accstd{76.98}{0.56}
& \accstd{89.21}{0.62} & \accstd{68.62}{0.21} & \accstd{93.62}{0.50} \\
Strong ceiling   
& \accstd{72.54}{0.65} & \accstd{92.64}{0.46} & \accstd{90.24}{0.33}
& \accstd{89.28}{0.77} & \accstd{64.88}{0.55} & \accstd{81.19}{0.72}
& \accstd{91.06}{0.47} & \accstd{70.23}{0.69} & \accstd{96.48}{0.72} \\
Naive            
& \accstd{73.11}{0.39} & \accstd{92.72}{0.72} & \accstd{88.49}{0.43}
& \accstd{90.83}{0.77} & \accstd{64.23}{0.72} & \accstd{79.24}{0.72}
& \accstd{90.24}{0.77} & \accstd{71.06}{0.33} & \accstd{95.66}{0.45} \\
Confidence-based 
& \accstd{73.28}{0.72} & \accstd{93.86}{0.65} & \accstd{88.18}{0.62}
& \accstd{91.00}{0.35} & \accstd{64.19}{0.50} & \accstd{79.45}{0.70}
& \accstd{87.87}{0.77} & \accstd{69.41}{0.35} & \accstd{93.94}{0.72} \\
SEAM             
& \accstd{63.36}{0.69} & \accstd{95.25}{0.68} & \accstd{87.98}{0.54}
& \accstd{93.86}{0.50} & \accstd{63.22}{0.77} & \accstd{83.25}{0.72}
& \accstd{84.07}{0.55} & \accstd{63.88}{0.72} & \accstd{93.37}{0.33} \\
\textsc{Anchor}  
& \accstd{73.41}{0.57} & \accstd{93.37}{0.21} & \accstd{89.00}{0.35}
& \accstd{91.41}{0.44} & \accstd{65.01}{0.72} & \accstd{81.60}{0.22}
& \accstd{90.65}{0.24} & \accstd{71.24}{0.54} & \accstd{95.34}{0.77} \\

\hdashline

\multicolumn{10}{l}{\emph{Qwen3-1.7B $\rightarrow$ Qwen3-8B}} \\
& \multicolumn{3}{c}{\textbf{Src: Anthropic Harmless}} 
& \multicolumn{3}{c}{\textbf{Src: PKU}} 
& \multicolumn{3}{c}{\textbf{Src: RAIL}} \\
\cmidrule(lr){2-4} \cmidrule(lr){5-7} \cmidrule(lr){8-10}
\textbf{Method} 
& AHarmless & PKU & RAIL
& PKU & AHarmless & RAIL
& RAIL & AHarmless & PKU \\
\midrule
Weak teacher     
& \accstd{71.80}{0.72} & \accstd{87.97}{0.28} & \accstd{82.73}{0.77}
& \accstd{87.97}{0.48} & \accstd{63.58}{0.77} & \accstd{81.50}{0.27}
& \accstd{89.31}{0.41} & \accstd{67.75}{0.56} & \accstd{91.24}{0.72} \\
Strong ceiling   
& \accstd{72.72}{0.33} & \accstd{93.21}{0.67} & \accstd{87.46}{0.53}
& \accstd{88.05}{0.42} & \accstd{64.53}{0.72} & \accstd{81.50}{0.77}
& \accstd{91.98}{0.72} & \accstd{69.58}{0.58} & \accstd{95.34}{0.77} \\
Naive            
& \accstd{73.28}{0.24} & \accstd{90.02}{0.77} & \accstd{83.25}{0.34}
& \accstd{88.79}{0.46} & \accstd{64.93}{0.46} & \accstd{83.35}{0.72}
& \accstd{88.80}{0.72} & \accstd{69.28}{0.72} & \accstd{92.96}{0.35} \\
Confidence-based 
& \accstd{71.85}{0.21} & \accstd{86.25}{0.45} & \accstd{82.22}{0.19}
& \accstd{89.20}{0.77} & \accstd{63.19}{0.77} & \accstd{83.04}{0.51}
& \accstd{88.49}{0.72} & \accstd{69.63}{0.30} & \accstd{95.17}{0.47} \\
SEAM             
& \accstd{57.09}{0.31} & \accstd{93.37}{0.72} & \accstd{81.19}{0.20}
& \accstd{91.33}{0.77} & \accstd{55.09}{0.72} & \accstd{78.83}{0.70}
& \accstd{79.03}{0.76} & \accstd{56.96}{0.19} & \accstd{92.31}{0.61} \\
\textsc{Anchor}  
& \accstd{73.28}{0.77} & \accstd{88.46}{0.53} & \accstd{84.38}{0.72}
& \accstd{90.18}{0.32} & \accstd{65.01}{0.67} & \accstd{82.43}{0.34}
& \accstd{89.31}{0.77} & \accstd{69.15}{0.71} & \accstd{93.70}{0.72} \\
\bottomrule
\end{tabular}
}
\\
\vspace{1mm}
\footnotesize{AHelpful = Anthropic Helpful, UltraFB = UltraFeedback, AHarmless = Anthropic Harmless}
\end{table*}

\subsection{Gap to Strong-Supervised Ceiling}

To further contextualize weak-to-strong performance, we compare each student with a strong reward model trained directly on gold labels from the evaluation domain. For a student trained on source domain $S$ and evaluated on domain $T$, we define the target-ceiling gap as

\[
    G_{S\rightarrow T} = A(m_T^{\mathrm{gt}},T) - A(m_S^{\mathrm{w2s}},T),
\]

where $m_T^{\mathrm{gt}}$ denotes the strong model trained directly on gold labels from $T$. When $T=S$, this reduces to the in-distribution strong-ceiling gap; when $T\neq S$, it measures how far a zero-shot source-trained student remains from direct target-domain supervision.

\begin{table}[ht]
\centering
\caption{Gap between weak-to-strong students and strong models trained directly on gold labels from the evaluation domain. ID and OOD columns report mean target-ceiling gaps in accuracy points. ``Weak-sup. gap'' denotes the additional OOD gap attributable to replacing gold source supervision with weak supervision. Lowest-gap counts use fractional counting for ties.}
\label{tab:ceiling_gap}
\small
\resizebox{\linewidth}{!}{%
\begin{tabular}{lccccc}
\toprule
Method &
ID gap $\downarrow$ &
OOD gap $\downarrow$ &
Weak-sup. gap $\downarrow$ &
Lowest ID &
Lowest OOD \\
\midrule
Naive W2S
    & 1.01 & 6.37 & 1.15 & 0.5 / 12 & 4 / 24 \\
Conf.-based
    & 1.56 & 6.27 & 1.05 & 0 / 12 & 2 / 24 \\
SEAM
    & 8.73 & 8.05 & 2.83 & 2 / 12 & 7 / 24 \\
\textsc{Anchor}
    & \textbf{-0.11} & \textbf{5.45} & \textbf{0.23}
    & \textbf{9.5 / 12} & \textbf{11 / 24} \\
\bottomrule
\end{tabular}
}
\end{table}

Table~\ref{tab:ceiling_gap} shows that \textsc{Anchor} has the smallest mean gap to the strong-supervised reference under both ID and OOD evaluation. Its mean ID gap is $-0.11$ points, slightly exceeding the source-domain strong-supervised reference on average, while its mean OOD target-ceiling gap is $5.45$ points, compared with $6.37$, $6.27$, and $8.05$ for Naive W2S, confidence-based W2S, and SEAM, respectively. \textsc{Anchor} also obtains the lowest gap in $9.5/12$ ID and $11/24$ OOD cases under fractional tie counting.

To distinguish limitations due to cross-domain transfer from those due specifically to weak supervision, we further decompose the OOD gap as

\[
\begin{array}{rcl}
G_{S\rightarrow T}
&=&
\underbrace{
A(m_T^{\mathrm{gt}},T)
-
A(m_S^{\mathrm{gt}},T)
}_{\text{source-target gold-transfer gap}}
\\[2pt]
&&+
\underbrace{
A(m_S^{\mathrm{gt}},T)
-
A(m_S^{\mathrm{w2s}},T)
}_{\text{weak-supervision gap}} .
\end{array}
\]

The first term measures the difficulty of transferring even gold source-domain supervision to the unseen target domain, whereas the second measures the additional gap introduced by weak-supervised training when the source dataset and strong-model architecture are held fixed. Averaged across OOD evaluations, the source-target gold-transfer gap is $5.22$ points. The additional weak-supervision gap is only $0.23$ points for \textsc{Anchor}, compared with $1.15$, $1.05$, and $2.83$ for Naive W2S, confidence-based W2S, and SEAM, respectively. \textsc{Anchor} substantially reduces the remaining discrepancy associated with weak-label training, while much of its residual OOD gap reflects source-target transfer difficulty that persists even under gold source-domain supervision.

%% file: content/appendix_materials/app_e.tex
\section{Anchoring Ablations and Representation Analysis}

\subsection{Expanded \texorpdfstring{$\lambda$}{lambda} and layer ablations}

The main-text ablations in Section~\ref{sec:ablation_studies} use Llama with HelpSteer3 as the source domain. Here, we extend both the anchoring-strength and layer-selection analyses to two additional Qwen settings: HelpSteer3 and UltraFeedback as source domains. These experiments test whether the conclusions from the main ablation persist across model families and source datasets.

\paragraph{Anchoring strength.}
Table~\ref{tab:expanded_lambda} reports the raw pairwise accuracies for $\lambda\in\{10^{-2},10^{-3},10^{-4}\}$. In addition to the Llama/HelpSteer3 setting reported in the main paper, we evaluate Qwen trained on UltraFeedback and HelpSteer3. Importantly, $\lambda=10^{-4}$ was selected using the original Llama/HelpSteer3 experiment and then fixed for the Qwen experiments rather than tuned independently for each source-target pair.

\begin{table}[ht]
\centering
\caption{Expanded ablation on the anchoring coefficient $\lambda$. We report pairwise accuracy (\%) in the evaluation order shown in the second column. H3, AH, and UF denote HelpSteer3, Anthropic Helpful, and UltraFeedback, respectively.}
\label{tab:expanded_lambda}
\small
\setlength{\tabcolsep}{4pt}
\resizebox{\linewidth}{!}{%
\begin{tabular}{llccc}
\toprule
Model / source & Eval. order &
$\lambda=10^{-2}$ &
$\lambda=10^{-3}$ &
$\lambda=10^{-4}$ \\
\midrule
Llama / H3
& H3 / AH / UF
& 76.61 / 58.21 / 70.22
& 76.72 / 60.06 / 70.58
& \textbf{77.60 / 61.15 / 71.82} \\

Qwen / UF
& UF / AH / H3
& 77.37 / 60.55 / 69.73
& 77.84 / 62.95 / 70.16
& \textbf{77.84 / 65.22 / 70.49} \\

Qwen / H3
& H3 / AH / UF
& 73.33 / 54.31 / 71.31
& 73.33 / 54.93 / 71.64
& \textbf{76.07 / 56.82 / 72.11} \\
\bottomrule
\end{tabular}
}
\end{table}

Across these settings, $\lambda=10^{-4}$ obtains the highest or tied-highest accuracy in all nine ID/OOD comparisons. Stronger anchoring generally reduces performance, suggesting that excessive preservation can restrict adaptation to weak preference supervision. At the same time, the difficult Qwen HelpSteer3$\rightarrow$Anthropic Helpful transfer remains weak even at $\lambda=10^{-4}$, indicating that this failure cannot be explained solely by overly strong anchoring.

\paragraph{Anchoring layer.}
We next compare final-layer ANCHOR with middle-layer anchoring at the fixed coefficient $\lambda=10^{-4}$. Table~\ref{tab:expanded_layer} extends the main-text layer ablation to the same two Qwen settings.

\begin{table}[ht]
\centering
\caption{Expanded comparison of middle- and final-layer anchoring at $\lambda=10^{-4}$. Values are pairwise accuracy (\%).}
\label{tab:expanded_layer}
\small
\setlength{\tabcolsep}{5pt}
\resizebox{\linewidth}{!}{%
\begin{tabular}{llcc}
\toprule
Model / source & Eval. order &
Middle layer & Final layer \\
\midrule
Llama / H3
& H3 / AH / UF
& 76.17 / 57.38 / \textbf{72.11}
& \textbf{77.60 / 61.15} / 71.82 \\

Qwen / UF
& UF / AH / H3
& 77.64 / 65.05 / 69.01
& \textbf{77.84 / 65.22 / 70.49} \\

Qwen / H3
& H3 / AH / UF
& 75.05 / 55.43 / 71.73
& \textbf{76.07 / 56.82 / 72.11} \\
\bottomrule
\end{tabular}
}
\end{table}

Final-layer anchoring achieves higher accuracy in eight of the nine comparisons, although several differences are small. Middle-layer anchoring is slightly better for Llama HelpSteer3$\rightarrow$ UltraFeedback, while final-layer anchoring provides the stronger overall ID/OOD balance across the evaluated model families and source domains. These results support our use of final-layer ANCHOR as the default variant, while also indicating that the optimal anchoring location can vary for individual source-target pairs.

\subsection{Representation Drift Analysis}
\label{app:representation_drift}

\begin{figure*}[t]
    \centering
    \includegraphics[width=0.9\textwidth]{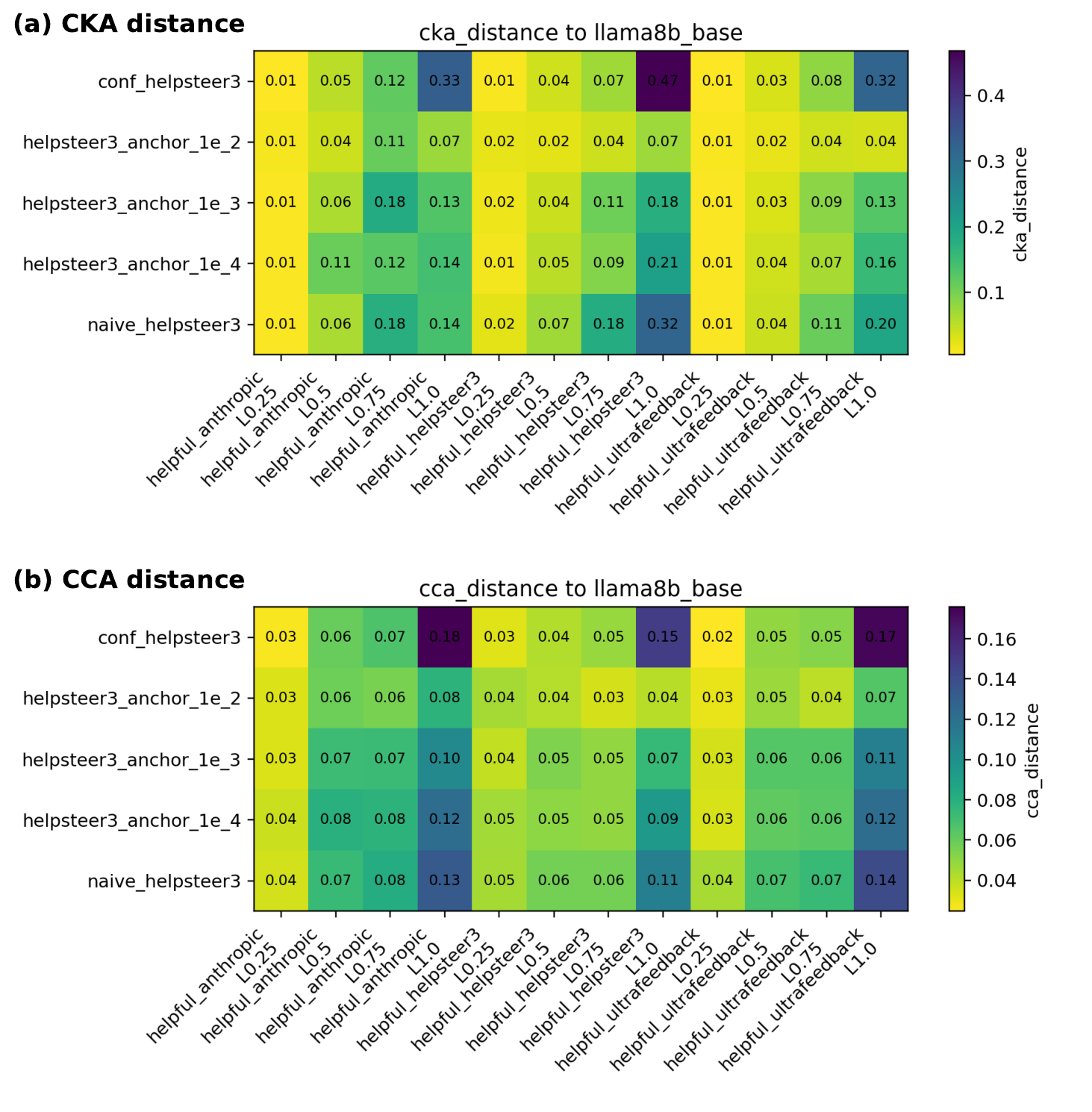}
    \caption{CKA and CCA distances between HelpSteer3-trained LoRA checkpoints and the frozen Llama-8B base model across evaluation domains and layers.}
    \label{fig:cka_cca_drift}
\end{figure*}

We further analyze how weak-supervised fine-tuning changes the internal representations of the strong model. Specifically, we compute CKA and CCA distances between each fine-tuned LoRA checkpoint and the frozen Llama-8B base model at different layers. Lower distance indicates that the fine-tuned model preserves representations closer to the pretrained base model, while higher distance indicates larger representation drift.

Figure~\ref{fig:cka_cca_drift} shows that representation drift is concentrated in middle and final layers, which is expected because these layers are closer to the reward head and are more directly affected by preference learning. Compared with naive and confidence-based weak-to-strong training, \textsc{Anchor} generally reduces CKA/CCA distance to the pretrained base model, indicating that anchoring constrains excessive representational movement during weak-label training.

Importantly, the best-performing anchored checkpoint is not the one with the smallest representation distance. Very strong anchoring, such as \(\lambda=10^{-2}\), keeps the representations closer to the base model, but can overly restrict adaptation to the source preference task. In contrast, the best-performing checkpoint, \(\lambda=10^{-4}\), allows moderate representation movement while still reducing drift relative to unanchored baselines. This supports a controlled-drift interpretation: \textsc{Anchor} improves weak-to-strong reward modeling not by freezing pretrained representations, but by balancing representation preservation with task-relevant adaptation.

\subsection{Preference-Relevant Linear Probing}

The CKA/CCA analysis measures geometric representation drift, but greater similarity to the pretrained representation space does not necessarily imply preservation of information relevant to preference prediction. We therefore conduct a complementary linear-probe analysis. We freeze the representations of Llama reward models trained on HelpSteer3 using Naive W2S, confidence-based W2S, and \textsc{Anchor}, and train the same linear probe using gold HelpSteer3 preference labels. We then evaluate the probe zero-shot on gold test labels from HelpSteer3, Anthropic Helpful, and UltraFeedback, without target-domain adaptation.

\begin{table}[ht]
\centering
\caption{Linear-probe accuracy (\%) on frozen representations of HelpSteer3-trained reward models. The probe is trained using gold HelpSteer3 preference labels and evaluated zero-shot on the three helpfulness datasets.}
\label{tab:linear_probe}
\small
\setlength{\tabcolsep}{6pt}
\resizebox{\linewidth}{!}{%
\begin{tabular}{lccc}
\toprule
Method & HelpSteer3 & Anthropic Helpful & UltraFeedback \\
\midrule
Naive W2S        & 71.07 & 55.06 & 69.33 \\
Conf.-based W2S  & 70.19 & 58.96 & 70.05 \\
\textsc{Anchor}  & \textbf{75.19} & \textbf{59.95} & \textbf{70.25} \\
\bottomrule
\end{tabular}
}
\end{table}

As shown in Table~\ref{tab:linear_probe}, \textsc{Anchor} achieves the highest probe accuracy on both the source dataset and the two unseen target datasets. In particular, it improves over Naive W2S by 4.12 points on HelpSteer3, 4.89 points on Anthropic Helpful, and 0.92 points on UltraFeedback. These results indicate that the representations preserved by \textsc{Anchor} contain more linearly accessible information predictive of gold preferences, complementing the geometric evidence from CKA/CCA. However, the probe captures aggregate preference-relevant information reflected in pairwise labels; it does not separately identify individual factors such as refusal behavior, verbosity, or other preference attributes.

%% file: content/appendix_materials/app_f.tex
\section{Preservation and Optimization Controls}
\label{sec:l2sp}

\subsection{Representation-Space versus Parameter-Space Preservation}
We additionally compare \textsc{Anchor} with L2-SP, a parameter-space regularization baseline that penalizes deviation from the pretrained initialization. The comparison is conducted when training on HelpSteer3 and evaluating zero-shot transfer to Anthropic Helpful and UltraFeedback, as reported in Table~\ref{tab:anchor_l2sp_helpsteer3}. The full sweep shows that L2-SP can improve transfer in some cases, especially on UltraFeedback, confirming that preserving pretrained information is useful. However, its effect is less balanced across target domains: the strongest L2-SP setting on UltraFeedback substantially underperforms \textsc{Anchor} on Anthropic Helpful and yields lower WRG. In contrast, \textsc{Anchor} achieves the best WRG and the strongest average transfer across the two target domains, suggesting that representation-space preservation provides a more stable mechanism than parameter-space regularization alone.

\begin{table}[t]
\centering
\small
\caption{Comparison between representation-space anchoring and parameter-space preservation when training on HelpSteer3. }
\label{tab:anchor_l2sp_helpsteer3}
\resizebox{\linewidth}{!}{%
\begin{tabular}{lccccc}
\toprule
\multirow{2}{*}{Method}
& \multirow{2}{*}{WRG}
& \multicolumn{2}{c}{Anthropic Helpful}
& \multicolumn{2}{c}{UltraFeedback} \\
\cmidrule(lr){3-4} \cmidrule(lr){5-6}
& & AOG & NTS & AOG & NTS \\
\midrule
\textsc{Anchor} $(10^{-2})$
& 2.58 & 5.25 & 5.25 & 6.33 & 6.33 \\
\textsc{Anchor} $(10^{-3})$
& \underline{2.69} & \underline{7.10} & \underline{7.10} & 6.69 & 6.69 \\
\textsc{Anchor} $(10^{-4})$
& \textbf{3.57} & \textbf{8.19} & \textbf{8.19} & \underline{7.93} & \underline{7.93} \\
\midrule
L2-SP $(10^{-3})$
& 2.30 & 5.63 & 5.63 & 7.33 & 7.33 \\
L2-SP $(10^{-4})$
& \underline{2.69} & 4.16 & 4.16 & \textbf{9.03} & \textbf{9.03} \\
L2-SP $(10^{-5})$
& 2.26 & 6.16 & 6.16 & 7.58 & 7.58 \\
\bottomrule
\end{tabular}
}
\end{table}

\subsection{Weight Interpolation and Optimization Controls}
\label{sec:optimization_controls}

We further test whether \textsc{Anchor}'s gains can be reproduced by simply limiting the magnitude, duration, or capacity of fine-tuning. We consider three controls: (1) weight interpolation (WI), which scales the Naive W2S LoRA update toward the pretrained initialization; (2) source-validation-based early stopping; and (3) reduced-capacity LoRA with rank $r=8$, compared with the default rank $r=16$. We evaluate these controls on both Llama and Qwen with HelpSteer3 as the source domain, and additionally evaluate WI on Llama/RAIL to cover the harmlessness category.

\begin{table}[t]
\centering
\caption{Comparison with weight interpolation and optimization/capacity controls. For HelpSteer3, Target 1 and Target 2 are Anthropic Helpful and UltraFeedback; for RAIL, they are Anthropic Harmless and PKU-SafeRLHF, respectively.}
\label{tab:optimization_controls}
\small
\setlength{\tabcolsep}{5pt}
\resizebox{\linewidth}{!}{%
\begin{tabular}{llrrrrr}
\toprule
Model / source & Method & WRG &
\multicolumn{2}{c}{Target 1} &
\multicolumn{2}{c}{Target 2} \\
\cmidrule(lr){4-5}\cmidrule(lr){6-7}
& & & AOG & NTS & AOG & NTS \\
\midrule

\multirow{7}{*}{Llama / HelpSteer3}
& WI ($\alpha=0.25$) & -22.01 & -0.30 & -22.31 & -15.39 & -37.40 \\
& WI ($\alpha=0.50$) & -17.20 &  3.73 & -13.47 & -10.01 & -27.21 \\
& WI ($\alpha=0.75$) &   1.38 &  4.12 &   4.12 &   5.32 &   5.32 \\
& WI ($\alpha=1.00$; Naive) & 1.71 & 1.46 & 1.46 & 4.80 & 4.80 \\
& Early stopping & 2.47 & 3.80 & 3.80 & 6.26 & 6.26 \\
& LoRA ($r=8$) & 1.49 & 2.86 & 2.86 & 6.14 & 6.14 \\
& \textsc{Anchor} & \textbf{3.57} & \textbf{8.19} & \textbf{8.19}
& \textbf{7.93} & \textbf{7.93} \\
\midrule

\multirow{7}{*}{Qwen / HelpSteer3}
& WI ($\alpha=0.25$) & -26.01 & -5.74 & -31.75 & -16.49 & -42.50 \\
& WI ($\alpha=0.50$) & -21.20 & -4.51 & -25.71 &  -8.51 & -29.71 \\
& WI ($\alpha=0.75$) &  -6.23 & -4.00 & -10.23 &  -0.12 &  -6.35 \\
& WI ($\alpha=1.00$; Naive) & -3.06 & -0.04 & -3.10 & 0.17 & -2.89 \\
& Early stopping & -3.84 & -3.17 & -7.01 & -0.58 & -4.42 \\
& LoRA ($r=8$) & -3.73 & -3.72 & -7.45 & -3.29 & -7.02 \\
& \textsc{Anchor} & \textbf{0.33} & \textbf{-2.61} & \textbf{-2.61}
& \textbf{4.69} & \textbf{4.69} \\
\midrule

\multirow{5}{*}{Llama / RAIL}
& WI ($\alpha=0.25$) & -31.35 & -13.62 & -44.97 & -33.15 & -64.50 \\
& WI ($\alpha=0.50$) &  -4.22 &  -2.48 &  -6.70 &  -1.89 &  -6.11 \\
& WI ($\alpha=0.75$) &   0.43 &   1.94 &   1.94 &   0.80 &   0.80 \\
& WI ($\alpha=1.00$; Naive) & 1.03 & 2.44 & 2.44 & 2.04 & 2.04 \\
& \textsc{Anchor} & \textbf{1.44} & \textbf{2.62} & \textbf{2.62}
& \textbf{2.37} & \textbf{2.37} \\
\bottomrule
\end{tabular}
}
\end{table}

Across these settings, the simpler controls do not reproduce \textsc{Anchor}'s overall ID/OOD balance, as depicted in Table~\ref{tab:optimization_controls}. Weight interpolation provides partial improvements for Llama/HelpSteer3 at $\alpha=0.75$, but smaller interpolation coefficients substantially degrade both source and target performance; the same pattern is more pronounced for Qwen/HelpSteer3 and Llama/RAIL. Source-validation early stopping and rank-$8$ LoRA also remain below \textsc{Anchor} in the HelpSteer3 experiments. In the difficult Qwen/HelpSteer3 setting, neither control restores positive source-domain gain, whereas \textsc{Anchor} achieves positive WRG and stronger transfer to UltraFeedback. Together with the L2-SP comparison above, these results indicate that \textsc{Anchor}'s gains are not explained solely by smaller parameter updates, shorter optimization, or reduced adaptation capacity.

%% file: custom.bib
@misc{amodei2016concreteproblemsaisafety,
      title={Concrete Problems in {AI} Safety}, 
      author={Dario Amodei and Chris Olah and Jacob Steinhardt and Paul Christiano and John Schulman and Dan Mané},
      year={2016},
      eprint={1606.06565},
      archivePrefix={arXiv},
      primaryClass={cs.AI},
      url={https://arxiv.org/abs/1606.06565}, 
}

@misc{leike2018scalableagentalignmentreward,
      title={Scalable agent alignment via reward modeling: a research direction}, 
      author={Jan Leike and David Krueger and Tom Everitt and Miljan Martic and Vishal Maini and Shane Legg},
      year={2018},
      eprint={1811.07871},
      archivePrefix={arXiv},
      primaryClass={cs.LG},
      url={https://arxiv.org/abs/1811.07871}, 
}

@misc{bowman2022measuringprogressscalableoversight,
      title={Measuring Progress on Scalable Oversight for Large Language Models}, 
      author={Samuel R. Bowman and Jeeyoon Hyun and Ethan Perez and Edwin Chen and Craig Pettit and Scott Heiner and Kamilė Lukošiūtė and Amanda Askell and Andy Jones and Anna Chen and Anna Goldie and Azalia Mirhoseini and Cameron McKinnon and Christopher Olah and Daniela Amodei and Dario Amodei and Dawn Drain and Dustin Li and Eli Tran-Johnson and Jackson Kernion and Jamie Kerr and Jared Mueller and Jeffrey Ladish and Joshua Landau and Kamal Ndousse and Liane Lovitt and Nelson Elhage and Nicholas Schiefer and Nicholas Joseph and Noemí Mercado and Nova DasSarma and Robin Larson and Sam McCandlish and Sandipan Kundu and Scott Johnston and Shauna Kravec and Sheer El Showk and Stanislav Fort and Timothy Telleen-Lawton and Tom Brown and Tom Henighan and Tristan Hume and Yuntao Bai and Zac Hatfield-Dodds and Ben Mann and Jared Kaplan},
      year={2022},
      eprint={2211.03540},
      archivePrefix={arXiv},
      primaryClass={cs.HC},
      url={https://arxiv.org/abs/2211.03540}, 
}

@article{bradley1952rank,
  author = {Bradley, Ralph Allan and Terry, Milton E.},
  title = {Rank Analysis of Incomplete Block Designs: The Method of Paired Comparisons},
  journal = {Biometrika},
  volume = {39},
  number = {3/4},
  pages = {324--345},
  year = {1952},
  doi = {10.1093/biomet/39.3-4.324},
}

@inproceedings{christiano2017deep,
  author = {Christiano, Paul F. and Leike, Jan and Brown, Tom B. and Martic, Miljan and Legg, Shane and Amodei, Dario},
  title = {Deep Reinforcement Learning from Human Preferences},
  booktitle = {Advances in Neural Information Processing Systems},
  volume = {30},
  year = {2017},
  url = {https://arxiv.org/abs/1706.03741},
}

@misc{ouyang2022training,
      title={Training language models to follow instructions with human feedback}, 
      author={Long Ouyang and Jeff Wu and Xu Jiang and Diogo Almeida and Carroll L. Wainwright and Pamela Mishkin and Chong Zhang and Sandhini Agarwal and Katarina Slama and Alex Ray and John Schulman and Jacob Hilton and Fraser Kelton and Luke Miller and Maddie Simens and Amanda Askell and Peter Welinder and Paul Christiano and Jan Leike and Ryan Lowe},
      year={2022},
      eprint={2203.02155},
      archivePrefix={arXiv},
      primaryClass={cs.CL},
      url={https://arxiv.org/abs/2203.02155}, 
}

@misc{bai2022training,
      title={Training a Helpful and Harmless Assistant with Reinforcement Learning from Human Feedback}, 
      author={Yuntao Bai and Andy Jones and Kamal Ndousse and Amanda Askell and Anna Chen and Nova DasSarma and Dawn Drain and Stanislav Fort and Deep Ganguli and Tom Henighan and Nicholas Joseph and Saurav Kadavath and Jackson Kernion and Tom Conerly and Sheer El-Showk and Nelson Elhage and Zac Hatfield-Dodds and Danny Hernandez and Tristan Hume and Scott Johnston and Shauna Kravec and Liane Lovitt and Neel Nanda and Catherine Olsson and Dario Amodei and Tom Brown and Jack Clark and Sam McCandlish and Chris Olah and Ben Mann and Jared Kaplan},
      year={2022},
      eprint={2204.05862},
      archivePrefix={arXiv},
      primaryClass={cs.CL},
      url={https://arxiv.org/abs/2204.05862}, 
}

@misc{burns2023weakstrong,
      title={Weak-to-Strong Generalization: Eliciting Strong Capabilities With Weak Supervision}, 
      author={Collin Burns and Pavel Izmailov and Jan Hendrik Kirchner and Bowen Baker and Leo Gao and Leopold Aschenbrenner and Yining Chen and Adrien Ecoffet and Manas Joglekar and Jan Leike and Ilya Sutskever and Jeff Wu},
      year={2023},
      eprint={2312.09390},
      archivePrefix={arXiv},
      primaryClass={cs.CL},
      url={https://arxiv.org/abs/2312.09390}, 
}

@inproceedings{yang2024weak,
    title = "Weak-to-Strong Reasoning",
    author = "Yang, Yuqing  and
      Ma, Yan  and
      Liu, Pengfei",
    editor = "Al-Onaizan, Yaser  and
      Bansal, Mohit  and
      Chen, Yun-Nung",
    booktitle = "Findings of the Association for Computational Linguistics: EMNLP 2024",
    month = nov,
    year = "2024",
    address = "Miami, Florida, USA",
    publisher = "Association for Computational Linguistics",
    url = "https://aclanthology.org/2024.findings-emnlp.490/",
    doi = "10.18653/v1/2024.findings-emnlp.490",
    pages = "8350--8367"
}

@inproceedings{li2024superfiltering,
    title = "Superfiltering: Weak-to-Strong Data Filtering for Fast Instruction-Tuning",
    author = "Li, Ming  and
      Zhang, Yong  and
      He, Shwai  and
      Li, Zhitao  and
      Zhao, Hongyu  and
      Wang, Jianzong  and
      Cheng, Ning  and
      Zhou, Tianyi",
    editor = "Ku, Lun-Wei  and
      Martins, Andre  and
      Srikumar, Vivek",
    booktitle = "Proceedings of the 62nd Annual Meeting of the Association for Computational Linguistics (Volume 1: Long Papers)",
    month = aug,
    year = "2024",
    address = "Bangkok, Thailand",
    publisher = "Association for Computational Linguistics",
    url = "https://aclanthology.org/2024.acl-long.769/",
    doi = "10.18653/v1/2024.acl-long.769",
    pages = "14255--14273"
}

@inproceedings{zhu2025weak,
    title={Weak-to-Strong Preference Optimization: Stealing Reward from Weak Aligned Model},
    author={Wenhong Zhu and Zhiwei He and Xiaofeng Wang and Pengfei Liu and Rui Wang},
    booktitle={The Thirteenth International Conference on Learning Representations},
    year={2025},
    url={https://openreview.net/forum?id=f7KxfUrRSb}
}

@inproceedings{cui2024ultrafeedback,
  title = 	 {{ULTRAFEEDBACK}: Boosting Language Models with Scaled {AI} Feedback},
  author =       {Cui, Ganqu and Yuan, Lifan and Ding, Ning and Yao, Guanming and He, Bingxiang and Zhu, Wei and Ni, Yuan and Xie, Guotong and Xie, Ruobing and Lin, Yankai and Liu, Zhiyuan and Sun, Maosong},
  booktitle = 	 {Proceedings of the 41st International Conference on Machine Learning},
  pages = 	 {9722--9744},
  year = 	 {2024},
  editor = 	 {Salakhutdinov, Ruslan and Kolter, Zico and Heller, Katherine and Weller, Adrian and Oliver, Nuria and Scarlett, Jonathan and Berkenkamp, Felix},
  volume = 	 {235},
  series = 	 {Proceedings of Machine Learning Research},
  month = 	 {21--27 Jul},
  publisher =    {PMLR},
  url = 	 {https://proceedings.mlr.press/v235/cui24f.html},
}

@misc{wang2025helpsteer3preference,
      title={{HelpSteer3-Preference}: Open Human-Annotated Preference Data across Diverse Tasks and Languages}, 
      author={Zhilin Wang and Jiaqi Zeng and Olivier Delalleau and Hoo-Chang Shin and Felipe Soares and Alexander Bukharin and Ellie Evans and Yi Dong and Oleksii Kuchaiev},
      year={2025},
      eprint={2505.11475},
      archivePrefix={arXiv},
      primaryClass={cs.CL},
      url={https://arxiv.org/abs/2505.11475}, 
}

@inproceedings{stiennon2020learning,
  title={Learning to summarize with human feedback},
  author={Stiennon, Nisan and Ouyang, Long and Wu, Jeffrey and Ziegler, Daniel and Lowe, Ryan and Voss, Chelsea and Radford, Alec and Amodei, Dario and Christiano, Paul F},
  booktitle={Advances in neural information processing systems},
  volume={33},
  pages={3008--3021},
  year={2020}
}

@inproceedings{rafailov2023direct,
  title={Direct preference optimization: Your language model is secretly a reward model},
  author={Rafailov, Rafael and Sharma, Archit and Mitchell, Eric and Manning, Christopher D and Ermon, Stefano and Finn, Chelsea},
  booktitle={Advances in neural information processing systems},
  volume={36},
  pages={53728--53741},
  year={2023}
}

@inproceedings{yang2024superficial,
    title={Super(ficial)-alignment: Strong Models May Deceive Weak Models in Weak-to-Strong Generalization},
    author={Wenkai Yang and Shiqi Shen and Guangyao Shen and Wei Yao and Yong Liu and Gong Zhi and Yankai Lin and Ji-Rong Wen},
    booktitle={The Thirteenth International Conference on Learning Representations},
    year={2025},
    url={https://openreview.net/forum?id=HxKSzulSD1}
}

@inproceedings{jiang2023llmblender,
    title = "{LLM}-Blender: Ensembling Large Language Models with Pairwise Ranking and Generative Fusion",
    author = "Jiang, Dongfu  and
      Ren, Xiang  and
      Lin, Bill Yuchen",
    editor = "Rogers, Anna  and
      Boyd-Graber, Jordan  and
      Okazaki, Naoaki",
    booktitle = "Proceedings of the 61st Annual Meeting of the Association for Computational Linguistics (Volume 1: Long Papers)",
    month = jul,
    year = "2023",
    address = "Toronto, Canada",
    publisher = "Association for Computational Linguistics",
    url = "https://aclanthology.org/2023.acl-long.792/",
    doi = "10.18653/v1/2023.acl-long.792",
    pages = "14165--14178",
}

@inproceedings{kumar2022finetuning,
    title={Fine-Tuning can Distort Pretrained Features and Underperform Out-of-Distribution},
    author={Ananya Kumar and Aditi Raghunathan and Robbie Matthew Jones and Tengyu Ma and Percy Liang},
    booktitle={International Conference on Learning Representations},
    year={2022},
    url={https://openreview.net/forum?id=UYneFzXSJWh}
}

@misc{hinton2015distilling,
  title={Distilling the Knowledge in a Neural Network}, 
  author={Geoffrey Hinton and Oriol Vinyals and Jeff Dean},
  year={2015},
  eprint={1503.02531},
  archivePrefix={arXiv},
  primaryClass={stat.ML},
  url={https://arxiv.org/abs/1503.02531}, 
}

@article{kirkpatrick2017overcoming,
    author = {James Kirkpatrick  and Razvan Pascanu  and Neil Rabinowitz  and Joel Veness  and Guillaume Desjardins  and Andrei A. Rusu  and Kieran Milan  and John Quan  and Tiago Ramalho  and Agnieszka Grabska-Barwinska  and Demis Hassabis  and Claudia Clopath  and Dharshan Kumaran  and Raia Hadsell },
    title = {Overcoming catastrophic forgetting in neural networks},
    journal = {Proceedings of the National Academy of Sciences},
    volume = {114},
    number = {13},
    pages = {3521-3526},
    year = {2017},
    doi = {10.1073/pnas.1611835114},
    URL = {https://www.pnas.org/doi/abs/10.1073/pnas.1611835114},
    eprint = {https://www.pnas.org/doi/pdf/10.1073/pnas.1611835114}
}

@article{li2017learning,
  title={Learning without forgetting},
  author={Li, Zhizhong and Hoiem, Derek},
  journal={IEEE transactions on pattern analysis and machine intelligence},
  volume={40},
  number={12},
  pages={2935--2947},
  year={2017},
  publisher={IEEE}
}

@inproceedings{charikar2024quantifying,
    title={Quantifying the Gain in Weak-to-Strong Generalization},
    author={Moses Charikar and Chirag Pabbaraju and Kirankumar Shiragur},
    booktitle={The Thirty-eighth Annual Conference on Neural Information Processing Systems},
    year={2024},
    url={https://openreview.net/forum?id=MyVyH5Jo1l}
}

@inproceedings{lee2020mixout,
    title={Mixout: Effective Regularization to Finetune Large-scale Pretrained Language Models},
    author={Cheolhyoung Lee and Kyunghyun Cho and Wanmo Kang},
    booktitle={International Conference on Learning Representations},
    year={2020},
    url={https://openreview.net/forum?id=HkgaETNtDB}
}

@misc{dubey2024llama,
      title={The {Llama 3} Herd of Models}, 
      author={Aaron Grattafiori and Abhimanyu Dubey and Abhinav Jauhri and Abhinav Pandey and Abhishek Kadian and Ahmad Al-Dahle and Aiesha Letman and Akhil Mathur and Alan Schelten and Alex Vaughan and Amy Yang and Angela Fan and Anirudh Goyal and Anthony Hartshorn and Aobo Yang and Archi Mitra and Archie Sravankumar and Artem Korenev and Arthur Hinsvark and Arun Rao and Aston Zhang and Aurelien Rodriguez and Austen Gregerson and Ava Spataru and Baptiste Roziere and Bethany Biron and Binh Tang and Bobbie Chern and Charlotte Caucheteux and Chaya Nayak and Chloe Bi and Chris Marra and Chris McConnell and Christian Keller and Christophe Touret and Chunyang Wu and Corinne Wong and Cristian Canton Ferrer and Cyrus Nikolaidis and Damien Allonsius and Daniel Song and Danielle Pintz and Danny Livshits and Danny Wyatt and David Esiobu and Dhruv Choudhary and Dhruv Mahajan and Diego Garcia-Olano and Diego Perino and Dieuwke Hupkes and Egor Lakomkin and Ehab AlBadawy and Elina Lobanova and Emily Dinan and Eric Michael Smith and Filip Radenovic and Francisco Guzmán and Frank Zhang and Gabriel Synnaeve and Gabrielle Lee and Georgia Lewis Anderson and Govind Thattai and Graeme Nail and Gregoire Mialon and Guan Pang and Guillem Cucurell and Hailey Nguyen and Hannah Korevaar and Hu Xu and Hugo Touvron and Iliyan Zarov and Imanol Arrieta Ibarra and Isabel Kloumann and Ishan Misra and Ivan Evtimov and Jack Zhang and Jade Copet and Jaewon Lee and Jan Geffert and Jana Vranes and Jason Park and Jay Mahadeokar and Jeet Shah and Jelmer van der Linde and Jennifer Billock and Jenny Hong and Jenya Lee and Jeremy Fu and Jianfeng Chi and Jianyu Huang and Jiawen Liu and Jie Wang and Jiecao Yu and Joanna Bitton and Joe Spisak and Jongsoo Park and Joseph Rocca and Joshua Johnstun and Joshua Saxe and Junteng Jia and Kalyan Vasuden Alwala and Karthik Prasad and Kartikeya Upasani and Kate Plawiak and Ke Li and Kenneth Heafield and Kevin Stone and Khalid El-Arini and Krithika Iyer and Kshitiz Malik and Kuenley Chiu and Kunal Bhalla and Kushal Lakhotia and Lauren Rantala-Yeary and Laurens van der Maaten and Lawrence Chen and Liang Tan and Liz Jenkins and Louis Martin and Lovish Madaan and Lubo Malo and Lukas Blecher and Lukas Landzaat and Luke de Oliveira and Madeline Muzzi and Mahesh Pasupuleti and Mannat Singh and Manohar Paluri and Marcin Kardas and Maria Tsimpoukelli and Mathew Oldham and Mathieu Rita and Maya Pavlova and Melanie Kambadur and Mike Lewis and Min Si and Mitesh Kumar Singh and Mona Hassan and Naman Goyal and Narjes Torabi and Nikolay Bashlykov and Nikolay Bogoychev and Niladri Chatterji and Ning Zhang and Olivier Duchenne and Onur Çelebi and Patrick Alrassy and Pengchuan Zhang and Pengwei Li and Petar Vasic and Peter Weng and Prajjwal Bhargava and Pratik Dubal and Praveen Krishnan and Punit Singh Koura and Puxin Xu and Qing He and Qingxiao Dong and Ragavan Srinivasan and Raj Ganapathy and Ramon Calderer and Ricardo Silveira Cabral and Robert Stojnic and Roberta Raileanu and Rohan Maheswari and Rohit Girdhar and Rohit Patel and Romain Sauvestre and Ronnie Polidoro and Roshan Sumbaly and Ross Taylor and Ruan Silva and Rui Hou and Rui Wang and Saghar Hosseini and Sahana Chennabasappa and Sanjay Singh and Sean Bell and Seohyun Sonia Kim and Sergey Edunov and Shaoliang Nie and Sharan Narang and Sharath Raparthy and Sheng Shen and Shengye Wan and Shruti Bhosale and Shun Zhang and Simon Vandenhende and Soumya Batra and Spencer Whitman and Sten Sootla and Stephane Collot and Suchin Gururangan and Sydney Borodinsky and Tamar Herman and Tara Fowler and Tarek Sheasha and Thomas Georgiou and Thomas Scialom and Tobias Speckbacher and Todor Mihaylov and Tong Xiao and Ujjwal Karn and Vedanuj Goswami and Vibhor Gupta and Vignesh Ramanathan and Viktor Kerkez and Vincent Gonguet and Virginie Do and Vish Vogeti and Vítor Albiero and Vladan Petrovic and Weiwei Chu and Wenhan Xiong and Wenyin Fu and Whitney Meers and Xavier Martinet and Xiaodong Wang and Xiaofang Wang and Xiaoqing Ellen Tan and Xide Xia and Xinfeng Xie and Xuchao Jia and Xuewei Wang and Yaelle Goldschlag and Yashesh Gaur and Yasmine Babaei and Yi Wen and Yiwen Song and Yuchen Zhang and Yue Li and Yuning Mao and Zacharie Delpierre Coudert and Zheng Yan and Zhengxing Chen and Zoe Papakipos and Aaditya Singh and Aayushi Srivastava and Abha Jain and Adam Kelsey and Adam Shajnfeld and Adithya Gangidi and Adolfo Victoria and Ahuva Goldstand and Ajay Menon and Ajay Sharma and Alex Boesenberg and Alexei Baevski and Allie Feinstein and Amanda Kallet and Amit Sangani and Amos Teo and Anam Yunus and Andrei Lupu and Andres Alvarado and Andrew Caples and Andrew Gu and Andrew Ho and Andrew Poulton and Andrew Ryan and Ankit Ramchandani and Annie Dong and Annie Franco and Anuj Goyal and Aparajita Saraf and Arkabandhu Chowdhury and Ashley Gabriel and Ashwin Bharambe and Assaf Eisenman and Azadeh Yazdan and Beau James and Ben Maurer and Benjamin Leonhardi and Bernie Huang and Beth Loyd and Beto De Paola and Bhargavi Paranjape and Bing Liu and Bo Wu and Boyu Ni and Braden Hancock and Bram Wasti and Brandon Spence and Brani Stojkovic and Brian Gamido and Britt Montalvo and Carl Parker and Carly Burton and Catalina Mejia and Ce Liu and Changhan Wang and Changkyu Kim and Chao Zhou and Chester Hu and Ching-Hsiang Chu and Chris Cai and Chris Tindal and Christoph Feichtenhofer and Cynthia Gao and Damon Civin and Dana Beaty and Daniel Kreymer and Daniel Li and David Adkins and David Xu and Davide Testuggine and Delia David and Devi Parikh and Diana Liskovich and Didem Foss and Dingkang Wang and Duc Le and Dustin Holland and Edward Dowling and Eissa Jamil and Elaine Montgomery and Eleonora Presani and Emily Hahn and Emily Wood and Eric-Tuan Le and Erik Brinkman and Esteban Arcaute and Evan Dunbar and Evan Smothers and Fei Sun and Felix Kreuk and Feng Tian and Filippos Kokkinos and Firat Ozgenel and Francesco Caggioni and Frank Kanayet and Frank Seide and Gabriela Medina Florez and Gabriella Schwarz and Gada Badeer and Georgia Swee and Gil Halpern and Grant Herman and Grigory Sizov and Guangyi and Zhang and Guna Lakshminarayanan and Hakan Inan and Hamid Shojanazeri and Han Zou and Hannah Wang and Hanwen Zha and Haroun Habeeb and Harrison Rudolph and Helen Suk and Henry Aspegren and Hunter Goldman and Hongyuan Zhan and Ibrahim Damlaj and Igor Molybog and Igor Tufanov and Ilias Leontiadis and Irina-Elena Veliche and Itai Gat and Jake Weissman and James Geboski and James Kohli and Janice Lam and Japhet Asher and Jean-Baptiste Gaya and Jeff Marcus and Jeff Tang and Jennifer Chan and Jenny Zhen and Jeremy Reizenstein and Jeremy Teboul and Jessica Zhong and Jian Jin and Jingyi Yang and Joe Cummings and Jon Carvill and Jon Shepard and Jonathan McPhie and Jonathan Torres and Josh Ginsburg and Junjie Wang and Kai Wu and Kam Hou U and Karan Saxena and Kartikay Khandelwal and Katayoun Zand and Kathy Matosich and Kaushik Veeraraghavan and Kelly Michelena and Keqian Li and Kiran Jagadeesh and Kun Huang and Kunal Chawla and Kyle Huang and Lailin Chen and Lakshya Garg and Lavender A and Leandro Silva and Lee Bell and Lei Zhang and Liangpeng Guo and Licheng Yu and Liron Moshkovich and Luca Wehrstedt and Madian Khabsa and Manav Avalani and Manish Bhatt and Martynas Mankus and Matan Hasson and Matthew Lennie and Matthias Reso and Maxim Groshev and Maxim Naumov and Maya Lathi and Meghan Keneally and Miao Liu and Michael L. Seltzer and Michal Valko and Michelle Restrepo and Mihir Patel and Mik Vyatskov and Mikayel Samvelyan and Mike Clark and Mike Macey and Mike Wang and Miquel Jubert Hermoso and Mo Metanat and Mohammad Rastegari and Munish Bansal and Nandhini Santhanam and Natascha Parks and Natasha White and Navyata Bawa and Nayan Singhal and Nick Egebo and Nicolas Usunier and Nikhil Mehta and Nikolay Pavlovich Laptev and Ning Dong and Norman Cheng and Oleg Chernoguz and Olivia Hart and Omkar Salpekar and Ozlem Kalinli and Parkin Kent and Parth Parekh and Paul Saab and Pavan Balaji and Pedro Rittner and Philip Bontrager and Pierre Roux and Piotr Dollar and Polina Zvyagina and Prashant Ratanchandani and Pritish Yuvraj and Qian Liang and Rachad Alao and Rachel Rodriguez and Rafi Ayub and Raghotham Murthy and Raghu Nayani and Rahul Mitra and Rangaprabhu Parthasarathy and Raymond Li and Rebekkah Hogan and Robin Battey and Rocky Wang and Russ Howes and Ruty Rinott and Sachin Mehta and Sachin Siby and Sai Jayesh Bondu and Samyak Datta and Sara Chugh and Sara Hunt and Sargun Dhillon and Sasha Sidorov and Satadru Pan and Saurabh Mahajan and Saurabh Verma and Seiji Yamamoto and Sharadh Ramaswamy and Shaun Lindsay and Shaun Lindsay and Sheng Feng and Shenghao Lin and Shengxin Cindy Zha and Shishir Patil and Shiva Shankar and Shuqiang Zhang and Shuqiang Zhang and Sinong Wang and Sneha Agarwal and Soji Sajuyigbe and Soumith Chintala and Stephanie Max and Stephen Chen and Steve Kehoe and Steve Satterfield and Sudarshan Govindaprasad and Sumit Gupta and Summer Deng and Sungmin Cho and Sunny Virk and Suraj Subramanian and Sy Choudhury and Sydney Goldman and Tal Remez and Tamar Glaser and Tamara Best and Thilo Koehler and Thomas Robinson and Tianhe Li and Tianjun Zhang and Tim Matthews and Timothy Chou and Tzook Shaked and Varun Vontimitta and Victoria Ajayi and Victoria Montanez and Vijai Mohan and Vinay Satish Kumar and Vishal Mangla and Vlad Ionescu and Vlad Poenaru and Vlad Tiberiu Mihailescu and Vladimir Ivanov and Wei Li and Wenchen Wang and Wenwen Jiang and Wes Bouaziz and Will Constable and Xiaocheng Tang and Xiaojian Wu and Xiaolan Wang and Xilun Wu and Xinbo Gao and Yaniv Kleinman and Yanjun Chen and Ye Hu and Ye Jia and Ye Qi and Yenda Li and Yilin Zhang and Ying Zhang and Yossi Adi and Youngjin Nam and Yu and Wang and Yu Zhao and Yuchen Hao and Yundi Qian and Yunlu Li and Yuzi He and Zach Rait and Zachary DeVito and Zef Rosnbrick and Zhaoduo Wen and Zhenyu Yang and Zhiwei Zhao and Zhiyu Ma},
      year={2024},
      eprint={2407.21783},
      archivePrefix={arXiv},
      primaryClass={cs.AI},
      url={https://arxiv.org/abs/2407.21783}, 
}

@misc{yang2025qwen3,
      title={Qwen3 Technical Report}, 
      author={An Yang and Anfeng Li and Baosong Yang and Beichen Zhang and Binyuan Hui and Bo Zheng and Bowen Yu and Chang Gao and Chengen Huang and Chenxu Lv and Chujie Zheng and Dayiheng Liu and Fan Zhou and Fei Huang and Feng Hu and Hao Ge and Haoran Wei and Huan Lin and Jialong Tang and Jian Yang and Jianhong Tu and Jianwei Zhang and Jianxin Yang and Jiaxi Yang and Jing Zhou and Jingren Zhou and Junyang Lin and Kai Dang and Keqin Bao and Kexin Yang and Le Yu and Lianghao Deng and Mei Li and Mingfeng Xue and Mingze Li and Pei Zhang and Peng Wang and Qin Zhu and Rui Men and Ruize Gao and Shixuan Liu and Shuang Luo and Tianhao Li and Tianyi Tang and Wenbiao Yin and Xingzhang Ren and Xinyu Wang and Xinyu Zhang and Xuancheng Ren and Yang Fan and Yang Su and Yichang Zhang and Yinger Zhang and Yu Wan and Yuqiong Liu and Zekun Wang and Zeyu Cui and Zhenru Zhang and Zhipeng Zhou and Zihan Qiu},
      year={2025},
      eprint={2505.09388},
      archivePrefix={arXiv},
      primaryClass={cs.CL},
      url={https://arxiv.org/abs/2505.09388}, 
}

@inproceedings{ji2024pku,
    title = "{PKU}-{S}afe{RLHF}: Towards Multi-Level Safety Alignment for {LLM}s with Human Preference",
    author = "Ji, Jiaming  and
      Hong, Donghai  and
      Zhang, Borong  and
      Chen, Boyuan  and
      Dai, Josef  and
      Zheng, Boren  and
      Qiu, Tianyi Alex  and
      Zhou, Jiayi  and
      Wang, Kaile  and
      Li, Boxun  and
      Han, Sirui  and
      Guo, Yike  and
      Yang, Yaodong",
    editor = "Che, Wanxiang  and
      Nabende, Joyce  and
      Shutova, Ekaterina  and
      Pilehvar, Mohammad Taher",
    booktitle = "Proceedings of the 63rd Annual Meeting of the Association for Computational Linguistics (Volume 1: Long Papers)",
    month = jul,
    year = "2025",
    address = "Vienna, Austria",
    publisher = "Association for Computational Linguistics",
    url = "https://aclanthology.org/2025.acl-long.1544/",
    doi = "10.18653/v1/2025.acl-long.1544",
    pages = "31983--32016",
    ISBN = "979-8-89176-251-0"
}

@misc{verma2025rail,
      title={{RAIL} in the Wild: Operationalizing Responsible {AI} Evaluation Using {Anthropic}'s Value Dataset}, 
      author={Sumit Verma and Pritam Prasun and Arpit Jaiswal and Pritish Kumar},
      year={2025},
      eprint={2505.00204},
      archivePrefix={arXiv},
      primaryClass={cs.AI},
      url={https://arxiv.org/abs/2505.00204}, 
}

@article{li2025strongempowered,
    title={Strong Empowered and Aligned Weak Mastered Annotation for Weak-to-Strong Generalization}, volume={39},
    url={https://ojs.aaai.org/index.php/AAAI/article/view/34955},
    DOI={10.1609/aaai.v39i26.34955}, number={26}, journal={Proceedings of the AAAI Conference on Artificial Intelligence}, author={Li, Yongqi and Miao, Xin and Xu, Mayi and Qian, Tieyun}, year={2025}, month={Apr.}, pages={27437–27445}
}

@inproceedings{shi2025mitigate,
    title = "How to Mitigate Overfitting in Weak-to-strong Generalization?",
    author = "Shi, Junhao  and
      Cheng, Qinyuan  and
      Fei, Zhaoye  and
      Zheng, Yining  and
      Guo, Qipeng  and
      Qiu, Xipeng",
    editor = "Che, Wanxiang  and
      Nabende, Joyce  and
      Shutova, Ekaterina  and
      Pilehvar, Mohammad Taher",
    booktitle = "Proceedings of the 63rd Annual Meeting of the Association for Computational Linguistics (Volume 1: Long Papers)",
    month = jul,
    year = "2025",
    address = "Vienna, Austria",
    publisher = "Association for Computational Linguistics",
    url = "https://aclanthology.org/2025.acl-long.784/",
    doi = "10.18653/v1/2025.acl-long.784",
    pages = "16100--16118",
    ISBN = "979-8-89176-251-0"
}

@inproceedings{nie2025weakforstrong,
    title={Weak-for-Strong:  Training Weak Meta-Agent to Harness Strong Executors},
    author={Fan Nie and Lan Feng and Haotian Ye and Weixin Liang and Pan Lu and Huaxiu Yao and Alexandre Alahi and James Zou},
    booktitle={Second Conference on Language Modeling},
    year={2025},
    url={https://openreview.net/forum?id=DmhcCRIfvq}
}

@inproceedings{lambert2025rewardbench,
    title = "{R}eward{B}ench: Evaluating Reward Models for Language Modeling",
    author = "Lambert, Nathan  and
      Pyatkin, Valentina  and
      Morrison, Jacob  and
      Miranda, LJ  and
      Lin, Bill Yuchen  and
      Chandu, Khyathi  and
      Dziri, Nouha  and
      Kumar, Sachin  and
      Zick, Tom  and
      Choi, Yejin  and
      Smith, Noah A.  and
      Hajishirzi, Hannaneh",
    editor = "Chiruzzo, Luis  and
      Ritter, Alan  and
      Wang, Lu",
    booktitle = "Findings of the Association for Computational Linguistics: NAACL 2025",
    month = apr,
    year = "2025",
    address = "Albuquerque, New Mexico",
    publisher = "Association for Computational Linguistics",
    url = "https://aclanthology.org/2025.findings-naacl.96/",
    doi = "10.18653/v1/2025.findings-naacl.96",
    pages = "1755--1797",
    ISBN = "979-8-89176-195-7"
}

@inproceedings{gheshlaghi2024general,
  title = 	 {A General Theoretical Paradigm to Understand Learning from Human Preferences},
  author =       {Gheshlaghi Azar, Mohammad and Daniel Guo, Zhaohan and Piot, Bilal and Munos, Remi and Rowland, Mark and Valko, Michal and Calandriello, Daniele},
  booktitle = 	 {Proceedings of The 27th International Conference on Artificial Intelligence and Statistics},
  pages = 	 {4447--4455},
  year = 	 {2024},
  editor = 	 {Dasgupta, Sanjoy and Mandt, Stephan and Li, Yingzhen},
  volume = 	 {238},
  series = 	 {Proceedings of Machine Learning Research},
  month = 	 {02--04 May},
  publisher =    {PMLR},
  url = 	 {https://proceedings.mlr.press/v238/gheshlaghi-azar24a.html},
}

@inproceedings{hong2024orpo,
    title = "{ORPO}: Monolithic Preference Optimization without Reference Model",
    author = "Hong, Jiwoo  and
      Lee, Noah  and
      Thorne, James",
    editor = "Al-Onaizan, Yaser  and
      Bansal, Mohit  and
      Chen, Yun-Nung",
    booktitle = "Proceedings of the 2024 Conference on Empirical Methods in Natural Language Processing",
    month = nov,
    year = "2024",
    address = "Miami, Florida, USA",
    publisher = "Association for Computational Linguistics",
    url = "https://aclanthology.org/2024.emnlp-main.626/",
    doi = "10.18653/v1/2024.emnlp-main.626",
    pages = "11170--11189"
}

@inproceedings{meng2024simpo,
    title={Sim{PO}: Simple Preference Optimization with a Reference-Free Reward},
    author={Yu Meng and Mengzhou Xia and Danqi Chen},
    booktitle={The Thirty-eighth Annual Conference on Neural Information Processing Systems},
    year={2024},
    url={https://openreview.net/forum?id=3Tzcot1LKb}
}

@inproceedings{li2024dual,
 author = {Li, Kaican and Xie, Weiyan and Huang, Yongxiang and Deng, Didan and Hong, Lanqing and Li, Zhenguo and Silva, Ricardo and Zhang, Nevin L.},
 booktitle = {Advances in Neural Information Processing Systems},
 doi = {10.52202/079017-2110},
 editor = {A. Globerson and L. Mackey and D. Belgrave and A. Fan and U. Paquet and J. Tomczak and C. Zhang},
 pages = {66025--66057},
 publisher = {Curran Associates, Inc.},
 title = {Dual Risk Minimization: Towards Next-Level Robustness in Fine-tuning Zero-Shot Models},
 url = {https://proceedings.neurips.cc/paper_files/paper/2024/file/7972f3735e104a54715922aa416fde1b-Paper-Conference.pdf},
 volume = {37},
 year = {2024}
}

@inproceedings{zang2024overcoming,
    title={Overcoming the Pitfalls of Vision-Language Model Finetuning for {OOD} Generalization},
    author={Yuhang Zang and Hanlin Goh and Joshua M. Susskind and Chen Huang},
    booktitle={The Twelfth International Conference on Learning Representations},
    year={2024},
    url={https://openreview.net/forum?id=PKICZXVY9M}
}

@inproceedings{li2024revisiting,
    title = "Revisiting Catastrophic Forgetting in Large Language Model Tuning",
    author = "Li, Hongyu  and
      Ding, Liang  and
      Fang, Meng  and
      Tao, Dacheng",
    editor = "Al-Onaizan, Yaser  and
      Bansal, Mohit  and
      Chen, Yun-Nung",
    booktitle = "Findings of the Association for Computational Linguistics: EMNLP 2024",
    month = nov,
    year = "2024",
    address = "Miami, Florida, USA",
    publisher = "Association for Computational Linguistics",
    url = "https://aclanthology.org/2024.findings-emnlp.249/",
    doi = "10.18653/v1/2024.findings-emnlp.249",
    pages = "4297--4308"
}

@inproceedings{zheng2025spurious,
    title={Spurious Forgetting in Continual Learning of Language Models},
    author={Junhao Zheng and Xidi Cai and Shengjie Qiu and Qianli Ma},
    booktitle={The Thirteenth International Conference on Learning Representations},
    year={2025},
    url={https://openreview.net/forum?id=ScI7IlKGdI}
}

@inproceedings{lyu2025macpo,
    title={{MACPO}: Weak-to-Strong Alignment via Multi-Agent Contrastive Preference Optimization},
    author={Yougang Lyu and Lingyong Yan and Zihan Wang and Dawei Yin and Pengjie Ren and Maarten de Rijke and Zhaochun Ren},
    booktitle={The Thirteenth International Conference on Learning Representations},
    year={2025},
    url={https://openreview.net/forum?id=x1Okv4kbVR}
}

@inproceedings{zhou2024weakstrongsearch,
    title={Weak-to-Strong Search: Align Large Language Models via Searching over Small Language Models},
    author={Zhanhui Zhou and Zhixuan Liu and Jie Liu and Zhichen Dong and Chao Yang and Yu Qiao},
    booktitle={The Thirty-eighth Annual Conference on Neural Information Processing Systems},
    year={2024},
    url={https://openreview.net/forum?id=dOJ6CqWDf1}
}

@inproceedings{liusie2024llm,
    title = "{LLM} Comparative Assessment: Zero-shot {NLG} Evaluation through Pairwise Comparisons using Large Language Models",
    author = "Liusie, Adian  and
      Manakul, Potsawee  and
      Gales, Mark",
    editor = "Graham, Yvette  and
      Purver, Matthew",
    booktitle = "Proceedings of the 18th Conference of the European Chapter of the Association for Computational Linguistics (Volume 1: Long Papers)",
    month = mar,
    year = "2024",
    address = "St. Julian{'}s, Malta",
    publisher = "Association for Computational Linguistics",
    url = "https://aclanthology.org/2024.eacl-long.8/",
    doi = "10.18653/v1/2024.eacl-long.8",
    pages = "139--151"
}

@inproceedings{li2024generative,
    title={Generative Judge for Evaluating Alignment},
    author={Junlong Li and Shichao Sun and Weizhe Yuan and Run-Ze Fan and Hai Zhao and Pengfei Liu},
    booktitle={The Twelfth International Conference on Learning Representations},
    year={2024},
    url={https://openreview.net/forum?id=gtkFw6sZGS}
}

@inproceedings{kim2024prometheus,
    title = "Prometheus 2: An Open Source Language Model Specialized in Evaluating Other Language Models",
    author = "Kim, Seungone  and
      Suk, Juyoung  and
      Longpre, Shayne  and
      Lin, Bill Yuchen  and
      Shin, Jamin  and
      Welleck, Sean  and
      Neubig, Graham  and
      Lee, Moontae  and
      Lee, Kyungjae  and
      Seo, Minjoon",
    editor = "Al-Onaizan, Yaser  and
      Bansal, Mohit  and
      Chen, Yun-Nung",
    booktitle = "Proceedings of the 2024 Conference on Empirical Methods in Natural Language Processing",
    month = nov,
    year = "2024",
    address = "Miami, Florida, USA",
    publisher = "Association for Computational Linguistics",
    url = "https://aclanthology.org/2024.emnlp-main.248/",
    doi = "10.18653/v1/2024.emnlp-main.248",
    pages = "4334--4353"
}

@inproceedings{wu2024meta,
    title={Meta Continual Learning Revisited: Implicitly Enhancing Online {Hessian} Approximation via Variance Reduction},
    author={Yichen Wu and Long-Kai Huang and Renzhen Wang and Deyu Meng and Ying Wei},
    booktitle={The Twelfth International Conference on Learning Representations},
    year={2024},
    url={https://openreview.net/forum?id=TpD2aG1h0D}
}
